\documentclass{article}

\PassOptionsToPackage{numbers, compress}{natbib}

\usepackage[final]{neurips_2023}

\usepackage[utf8]{inputenc} % allow utf-8 input
\usepackage[T1]{fontenc}    % use 8-bit T1 fonts
\usepackage{graphicx}
\usepackage{amsmath,amssymb,amsthm,mathabx}
\usepackage{booktabs}
\usepackage{algorithmic}
\usepackage[linesnumbered,ruled,vlined]{algorithm2e}
\usepackage{acronym}
\usepackage{enumitem}
\usepackage[
    colorlinks, 
    linkcolor=green,
    anchorcolor=blue, 
    citecolor=blue
]{hyperref}
\usepackage{setspace}
\usepackage{subfigure}
\usepackage[skip=3pt,font=small]{caption}
\usepackage[dvipsnames,table]{xcolor}
\usepackage[capitalise]{cleveref}
\usepackage{tabularx,multirow,array,makecell}
\usepackage{overpic,wrapfig}

\usepackage{listings}
\lstdefinestyle{mystyle}{
  backgroundcolor=\color{backcolour},   commentstyle=\color{codegreen},
  keywordstyle=\color{magenta},
  numberstyle=\tiny\color{codegray},
  stringstyle=\color{codepurple},
  basicstyle=\ttfamily\footnotesize,
  commentstyle=\color{red!10!green!70}\textit,
  breakatwhitespace=false,         
  breaklines=true,                 
  captionpos=b,                    
  keepspaces=true,                 
  numbers=left,                    
  numbersep=5pt,                  
  showspaces=false,                
  showstringspaces=false,
  showtabs=false,                  
  tabsize=2
}

\usepackage{tikz}
\usetikzlibrary{positioning, arrows.meta, quotes}
\usetikzlibrary{shapes,decorations}
\usetikzlibrary{bayesnet}
\tikzset{>=latex}
\tikzstyle{plate caption} = [
    caption, node distance=0, inner sep=0pt,
    below left=5pt and 0pt of #1.south
]
\tikzstyle{nbase} = [
    circle, 
    draw=black, 
    fill=white, 
    inner sep=0pt, 
    minimum size=1.0cm,
    node distance=16mm
]

\definecolor{gray}{gray}{0.9}
\definecolor{mgray}{gray}{0.7}
\definecolor{tgray}{gray}{0.5}

\makeatletter
\DeclareRobustCommand\onedot{\futurelet\@let@token\@onedot}
\def\@onedot{\ifx\@let@token.\else.\null\fi\xspace}
\def\eg{\emph{e.g}\onedot} 

\def\ie{\emph{i.e}\onedot}

\makeatother

\DeclareMathOperator*{\argmin}{arg\,min}

\newcommand{\w}{\boldsymbol{w}}
\newcommand{\x}{\boldsymbol{x}}

\newcommand{\z}{\boldsymbol{z}}
\newcommand{\E}{\mathbb{E}}
\newcommand{\R}{\mathbb{R}}
\newcommand{\I}{\mathbf{I}}
\newcommand{\N}{\mathcal{N}}

\newcommand{\mQ}{\mathcal{Q}}
\newcommand{\mK}{\mathcal{K}}
\newcommand{\mD}{\mathcal{D}}
\newcommand{\mH}{\mathcal{H}}
\newcommand{\mL}{\mathcal{L}}
\newcommand{\mI}{\mathcal{I}}
\newcommand{\bA}{\boldsymbol{\alpha}}
\newcommand{\bB}{\boldsymbol{\beta}}
\newcommand{\bT}{\boldsymbol{\theta}}
\newcommand{\bP}{\boldsymbol{\phi}}
\newcommand{\be}{\boldsymbol{\epsilon}}
\newcommand{\bM}{\boldsymbol{\mu}}
\newcommand{\bZ}{\boldsymbol{0}}

\makeatletter
\renewcommand{\paragraph}{%
  \@startsection{paragraph}{4}%
  {\z@}{0ex \@plus 0ex \@minus 0ex}{-1em}%
  {\hskip\parindent\normalfont\normalsize\bfseries}%
}
\makeatother

\graphicspath{{figures/}}

\crefname{algocf}{alg.}{algs.}
\Crefname{algocf}{Algrithm}{Algrithm}

\definecolor{gblue}{HTML}{4285F4}
\definecolor{gred}{HTML}{DB4437}

\acrodef{dlvm}[DLVM]{Deep Latent Variable Model}
\acrodef{vae}[VAE]{Variational Auto-Encoder}
\acrodef{ebm}[EBM]{Energy-Based Model}
\acrodef{lebm}[LEBM]{Latent-space Energy-Based Model}
\acrodef{mle}[MLE]{Maximum Likelihood Estimation}
\acrodef{ld}[LD]{Langevin Dynamics}
\acrodef{kld}[KLD]{Kullbeck-Leibler Divergence}
\acrodef{damc}[DAMC]{Diffusion-Amortized MCMC}
\acrodef{ddpm}[DDPM]{Denoising Diffusion Probabilistic Model}
\acrodef{abp}[ABP]{Alternating Back-Propogation}
\acrodef{sri}[SRI]{Short-Run Inference}
\acrodef{mcmc}[MCMC]{Markov Chain Monte Carlo}
\acrodef{pcd}[PCD]{Persistent Contrastive Divergence}
\acrodef{cd}[CD]{Contrastive Divergence}

\newcolumntype{C}{>{\columncolor{gray}}c}
\newcolumntype{L}{>{\columncolor{gray}}l}
\newcommand{\tabincell}[2]{\begin{tabular}{@{}#1@{}}#2\end{tabular}}

\title{Learning Energy-Based Prior Model with Diffusion-Amortized MCMC}

\author{
    Peiyu Yu$^1$ \\
    \texttt{yupeiyu98@g.ucla.edu}
    \And Yaxuan Zhu$^1$ \\
    \texttt{yaxuanzhu@g.ucla.edu}
    \And Sirui Xie$^2$ \\
    \texttt{srxie@ucla.edu}
    \And Xiaojian Ma$^{2,4}$ \\
    \texttt{xiaojian.ma@ucla.edu}
    \And Ruiqi Gao$^{3}$ \\
    \texttt{ruiqig@google.com}  
    \And Song-Chun Zhu$^{4}$ \\
    \texttt{sczhu@stat.ucla.edu}
    \And Ying Nian Wu$^1$ \\
    \texttt{ywu@stat.ucla.edu}
    \AND
    \normalfont{$^1$UCLA Department of Statistics\quad{}}
    \normalfont{$^2$UCLA Department of Computer Science}\\
    $^3$Google DeepMind
    $^4$Beijing Institute for General Artificial Intelligence (BIGAI)
}

\begin{document}

\maketitle

\begin{abstract}
Latent space \acp{ebm}, also known as energy-based priors, have drawn growing interests in the field of generative modeling due to its flexibility in the formulation and strong modeling power of the latent space. However, the common practice of learning latent space \acp{ebm} with non-convergent short-run \acs{mcmc} for prior and posterior sampling is hindering the model from further progress; the degenerate \acs{mcmc} sampling quality in practice often leads to degraded generation quality and instability in training, especially with highly multi-modal and/or high-dimensional target distributions. To remedy this sampling issue, in this paper we introduce a simple but effective diffusion-based amortization method for long-run \acs{mcmc} sampling and develop a novel learning algorithm for the latent space \ac{ebm} based on it. We provide theoretical evidence that the learned amortization of \acs{mcmc} is a valid long-run \acs{mcmc} sampler. Experiments on several image modeling benchmark datasets demonstrate the superior performance of our method compared with strong counterparts\footnote{Code and data available at \href{https://github.com/yuPeiyu98/Diffusion-Amortized-MCMC}{https://github.com/yuPeiyu98/Diffusion-Amortized-MCMC}.}.
\end{abstract}

\section{Introduction}
\label{sec:intro}
Generative modeling of data distributions has achieved impressive progress with the fast development of deep generative models in recent years \cite{kingma2013auto,rezende2014stochastic,goodfellow2014generative,rezende2015variational,sohl2015deep,razavi2019generating,song2019generative,ho2020denoising,song2020score}. It provides a powerful framework that allows successful applications in synthesizing data of different modalities \cite{karras2019style,kong2020diffwave,yu2021unsupervised,luo2021diffusion,vahdat2021score,yu2022latent}, extracting semantically meaningful data representation \cite{zhu2016generative,van2017neural,zhu2020domain} as well as other important domains of unsupervised or semi-supervised learning \citep{du2020compositional,grathwohl2019your,nie2021controllable}. A fundamental and powerful branch of generative modeling is the \ac{dlvm}. Typically, \ac{dlvm} assumes that the observation (\eg, a piece of text or images) is generated by its corresponding low-dimensional latent variables via a top-down generator network \cite{kingma2013auto,rezende2014stochastic,goodfellow2014generative}. The latent variables are often assumed to follow a non-informative prior distribution, such as a uniform or isotropic Gaussian distribution. While one can directly learn a deep top-down generator network to faithfully map the non-informative prior distribution to the data distribution, learning an informative prior model in the latent space could further improve the expressive power of the \ac{dlvm} with significantly less parameters \cite{pang2020learning}. In this paper, we specifically consider learning an \ac{ebm} in the latent space as an informative prior for the model. 

Learning energy-based prior can be challenging, as it typically requires computationally expensive \ac{mcmc} sampling to estimate learning gradients. The difficulty of MCMC-based sampling is non-negligible when the target distribution is highly multi-modal or high-dimensional. In these situations, MCMC sampling can take a long time to converge and perform poorly on traversing modes with limited iterations  \cite{nijkamp2020anatomy}. Consequently, training models with samples from non-convergent short-run MCMC \cite{nijkamp2019learning}, which is a common choice for learning latent space \acp{ebm} \cite{pang2020learning}, often results in malformed energy landscapes \cite{ yu2022latent,nijkamp2019learning,gao2020learning} and biased estimation of the model parameter. One possible solution is to follow the variational learning scheme \cite{kingma2013auto}, which however requires non-trivial extra efforts on model design to deal with problems like posterior collapse \cite{bowman2016generating, razavi2019preventing,lucas2019don} and limited expressivity induced by model assumptions \cite{kingma2013auto,ranganath2016hierarchical,tomczak2018vae}.

To remedy this sampling issue and further unleash the expressive power of the prior model, we propose to shift attention to an economical compromise between unrealistically expensive long-run MCMC and biased short-run MCMC: \textit{we consider learning valid amortization of the potentially long-run MCMC for learning energy-based priors}. Specifically, inspired by the connection between \ac{mcmc} sampling and denoising diffusion process \cite{song2019generative,ho2020denoising,song2020improved}, in this paper we propose a diffusion-based amortization method suitable for long-run \ac{mcmc} sampling in learning latent space \acp{ebm}. The learning algorithm derived from it breaks the long-run chain into consecutive affordable short-run segments that can be iteratively distilled by a diffusion-based sampler. The core idea is simple and can be summarized by a one-liner (\cref{fig:method}). We provide theoretical and empirical evidence that the resulting sampler approximates the long-run chain (see the proof-of-concept toy examples in \cref{app:toy}), and brings significant performance improvement for learning latent space \acp{ebm} on several tasks. We believe that this proposal is a notable attempt to address the learning issues of energy-based priors and is new to the best of our knowledge. We kindly refer to \cref{sec:relatd_work} for a comprehensive discussion of the related work.

\paragraph{Contributions} i) We propose a diffusion-based amortization method for \ac{mcmc} sampling and develop a novel learning algorithm for the latent space \ac{ebm}. ii) We provide a theoretical understanding that the learned amortization of \ac{mcmc} is a valid long-run \ac{mcmc} sampler. iii) Our experiments demonstrate empirically that the proposed method brings higher sampling quality, a better-learned model and stronger performance on several image modeling benchmark datasets. 

\section{Background}
\label{sec:background}
\subsection{Energy-Based Prior Model}
\label{sec:lebm}
We assume that for the observed sample $\x \in \R^D$, there exists $\z \in \R^d$ as its unobserved latent variable vector. The complete-data distribution is
\begin{equation}
\label{equ:lebm_def}
    p_{\bT}(\z,\x) := p_{\bA}(\z)p_{\bB}(\x|\z),~
    p_{\bA}(\z) := \frac{1}{Z_{{\bA}}}\exp
                    \left(
                        f_{\bA}(\z)
                    \right)p_0(\z),
\end{equation}
where $p_{\bA}(\z)$ is the prior model with parameters ${\bA}$, $p_{\bB}(\x|\z)$ is the top-down generation model with parameters ${\bB}$, and ${\bT} = ({\bA},{\bB})$. The prior model $p_{\bA}(\z)$ can be formulated as an energy-based model, which we refer to as the \ac{lebm}~\cite{pang2020learning} throughout the paper. In this formulation, $f_{\bA}(\z)$ is parameterized by a neural network with scalar output, $Z_{\bA}$ is the partition function, and $p_0(\z)$ is standard normal as a reference distribution. The prior model in \cref{equ:lebm_def} can be interpreted as an energy-based correction or exponential tilting of the original prior distribution $p_0$. The generation model follows $p_{\bB}(\x|\z) = \N(g_{\bB}(\z), \sigma^2 \I_D)$, where $g_{\bB}$ is the generator network and $\sigma^2$ takes a pre-specified value as in \acs{vae} \cite{kingma2013auto}. This is equivalent to using $l_2$ error for reconstruction.

The parameters of \ac{lebm} and the generation model can be learned by \ac{mle} \cite{pang2020learning}. 
To be specific, given the training data $\x$, the gradients for updating ${\bA}, {\bB}$ are,
\begin{equation}
\label{equ:mle_grad}
\delta_{\bA} (\x)
    := \E_{p_{\bT}(\z|\x)} 
        \left[\nabla_{\bA} f_{\bA}(\z) \right] 
    - \E_{p_{\bA}(\z)} 
        \left[\nabla_{\bA} f_{\bA}(\z) \right],~
\delta_{\bB} (\x)
    := \E_{p_{\bT}(\z|\x)} 
        \left[\nabla_{\bB} \log p_{\bB}(\x | \z) \right].
\end{equation}

In practice, one may use the Monte-Carlo average to estimate the expectations in \cref{equ:mle_grad}. This involves sampling from the prior $p_{\bA}(\z)$ and the posterior $p_{\bT}(\z|\x)$ distribution using \ac{mcmc}, specifically \ac{ld} \cite{welling2011bayesian}, to estimate the expectations and hence the gradient. For a target distribution $\pi(\z)$, the dynamics iterates 
\begin{equation}
\label{equ:ld}
\z_{t + 1} = \z_t + \frac{s^2}{2} \nabla_{\z_t} \log \pi(\z_t) + s \w_t,~t=0,1,...,T-1,~\w_t \sim \N({\bZ}, \I_d),
\end{equation}
where $s$ is a small step size. One can draw $\z_0 \sim \N({\bZ}, \I_d)$ to initialize the chain. For sufficiently small step size $s$, the distribution of $\z_t$ will converge to $\pi$ as $t \to \infty$ \cite{welling2011bayesian}. However, it is prohibitively expensive to run \ac{ld} until convergence in most cases, for which we may resort to limited iterations of \ac{ld} for sampling in practice. This non-convergent short chain yields a moment-matching distribution close to the true $\pi(\z)$ but is often biased, which was dubbed as short-run \ac{ld} \cite{yu2022latent,nijkamp2020anatomy,nijkamp2019learning,gao2020learning}.

\subsection{Denoising Diffusion Probabilistic Model}
\label{sec:ddpm}
Closely related to \acp{ebm} are the \acp{ddpm} \cite{sohl2015deep,song2019generative,ho2020denoising}. As pointed out in \cite{sohl2015deep,ho2020denoising}, the sampling procedure of \ac{ddpm} with $\be$-prediction parametrization resembles \ac{ld}; $\be$ (predicted noise) plays a similar role to the gradient of the log density \cite{ho2020denoising}. 

In the formulation proposed by \citet{kingma2021variational}, the \ac{ddpm} parameterized by ${\bP}$ is specified by a noise schedule built upon $\lambda_s = \log[\beta_s^2/\sigma_s^2]$, \ie, the log signal-to-noise-ratio, that decreases monotonically with $s$. $\beta_s$ and $\sigma_s^2$ are strictly positive scalar-valued functions of $s$. We use $\z_0$ to denote training data in $\R^d$. The forward-time diffusion process $q(\z|\z_0)$ is defined as:
\begin{equation}
q(\z_s|\z_0) = \N(\z_s; \beta_s \z_0, \sigma_s^2 \I_d), \quad
q(\z_s'|\z_s) = \N(\z_s'; (\beta_{s'}/\beta_s)\z_s, \sigma_{s'|s}^2\I_d),
\end{equation}
where $0 \leq s < s' \leq S$ and $\sigma^2_{s'|s} = (1-e^{\lambda_{s'}-\lambda_s})\sigma_s^2$. Noticing that the forward process can be reverted as $q(\z_s|\z_{s'},\z_0) = \N(\z_s; \tilde{\bM}_{s|s'}(\z_{s'},\z_0), \tilde\sigma^2_{s|{s'}}\I_d)$, an ancestral sampler $q_{\bP}(\z_s|\z_{s'})$ \cite{ho2020denoising} that starts at $\z_S \sim \mathcal{N}({\bZ}, \I_d)$ can be derived accordingly \cite{kingma2021variational}:
\begin{equation}
\label{equ:ac_sampler}
\begin{aligned}
\tilde{\bM}_{s|s'}(\z_{s'},\z_0) &= e^{\lambda_{s'}-\lambda_s}(\alpha_s/\alpha_{s'})\z_{s'} + (1-e^{\lambda_{s'}-\lambda_s})\alpha_s\z_0, \quad
\tilde\sigma^2_{s|s'} = (1-e^{\lambda_{s'}-\lambda_s})\sigma_s^2, \\
\z_s &= \tilde{\bM}_{s|s'}(\z_{s'}, \hat\z_0) + \sqrt{(\tilde\sigma^2_{s|s'})^{1-\gamma} (\sigma^2_{s'|s})^\gamma)}{\be},
\end{aligned}
\end{equation}
where $\be$ is standard Gaussian noise, $\hat\z_{0}$ is the prediction of $\z_0$ by the \ac{ddpm} ${\bP}$, and $\gamma$ is a hyperparameter that controls the noise magnitude, following~\cite{nichol2021improved}. 
The goal of \ac{ddpm} is to recover the distribution of $\z_0$ from the given Gaussian noise distribution. It can be trained by optimizing
$
\E_{{\be},\lambda}\left[\|{\be}(\z_\lambda) - {\be} \|_2^2\right],
$
where ${\be} \sim \N({\bZ}, \I_d)$ and $\lambda$ is drawn from a distribution of log noise-to-signal ratio $p(\lambda)$
over uniformly sampled times $s \in [0,S]$. This loss can be justified as a lower bound on the data log-likelihood \cite{ho2020denoising,kingma2021variational} or as a variant of denoising score matching \cite{song2019generative, vincent2011connection}. We will exploit in this paper the connection between \acp{ddpm} and \ac{ld} sampling of \acp{ebm}, based upon which we achieve better sampling performance for \ac{lebm} compared with short-run \ac{ld}.

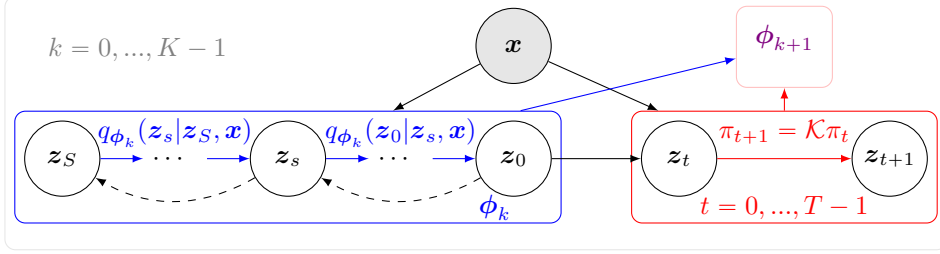
\begin{figure}[t!]
    \centering
    \begin{tikzpicture}
        %%%%% nodes
        \node [nbase] (zt) at (0.45,0) {$\z_{t}$};
        \node [nbase] (zt1) at (3.25,0) {$\z_{t+1}$};
        
        \node [nbase] (zT) at (-7.75,0) {$\z_S$};
        \node [nbase] (zs) at (-4.75,0) {$\z_s$};
        \node [nbase] (z0) at (-1.75,0) {$\z_0$};
        \node[text width=0.6cm] (ukn1) at (-6.25,0) {$\LARGE\cdots$};
        \node[text width=0.6cm] (ukn2) at (-3.25,0) {$\LARGE\cdots$};

        \node [nbase, fill=gray] (x) at (-1.75,1.5) {$\x$};
    
        %%%%% text
        \node (rev1) at (-6.25, 0.35) 
            {\color{blue}$q_{{\bP}_k}(\z_{s}|\z_S, \x)$};
        \node (rev2) at (-3.25, 0.35) 
            {\color{blue}$q_{{\bP}_k}(\z_{0}|\z_s, \x)$};
        \node (bpk) at (-2.0, -0.65)
            {\color{blue}${\bP}_k$};

        \node (K) at (1.85, 0.35)
            {\color{red}$\pi_{t+1} = {\mK} \pi_t$};
        \node (T_cond) at (1.85, -0.65)
            {\color{red}$t=0,...,T-1$};

        \node (bpk1) at (1.85, 1.6)
            {\color{blue!50!red}${\bP}_{k+1}$};

        \node (K_cond) at (-6.75, 1.45)
            {\color{tgray}$k=0,...,K-1$};
    
        %%%%% bbox    
        \plate [color=blue] {part1} {(z0)(zs)(zT)(ukn1)(ukn2)} {};
        \plate [color=red] {part2} {(zt)(zt1)} {};
        \plate [color=pink] {part3} {(bpk1)} {};
        \plate [color=gray] {all} {(part1)(part2)(part3)} {};
        
        %%%%% arrow
        \path [draw, ->, color=blue] (zT) edge (ukn1);
        \path [draw, ->, color=blue] (ukn1) edge (zs);
        \path [draw, ->, color=blue] (zs) edge (ukn2);
        \path [draw, ->, color=blue] (ukn2) edge (z0);
        \path [draw, ->, dashed] 
            (zs) edge [bend left] node [right] {} (zT);
        \path [draw, ->, dashed] 
            (z0) edge [bend left] node [right] {} (zs);

        \path [draw, ->] (z0) edge (zt);
        
        \path [draw, ->, color=red] (zt) edge (zt1);

        \path [draw, ->] (x) edge (part1);
        \path [draw, ->] (x) edge (part2);

        \path [draw, ->, color=blue] 
            (part1) edge (part3);
        \path [draw, ->, color=red] 
            (part2) edge (part3);
    \end{tikzpicture}
    \caption{\textbf{Learning the \acs{damc} sampler.} The training samples for updating the sampler to {\color{blue!50!red}${\bP}_{k+1}$} is obtained by {\color{red}$T$-step short-run \acs{ld}}, initialized with the samples from the current learned sampler {\color{blue}${\bP}_{k}$}. Best viewed in color.}
    \label{fig:method}
    \vspace{-10pt}
\end{figure}

\section{Method}
In this section, we introduce the diffusion-based amortization method for long-run \ac{mcmc} sampling in learning \ac{lebm} in \cref{sec:damc}. The learning algorithm of \ac{lebm} based on the proposed method and details about implementation are then presented in \cref{sec:app_mle} and \cref{sec:implem}, respectively.

\subsection{Amortizing MCMC with \acs{ddpm}}
\label{sec:damc}
\paragraph{Amortized MCMC}
Using the same notation as in \cref{sec:background}, we denote the starting distribution of \ac{ld} as $\pi_0(\z)$, the distribution after $t$-th iteration as $\pi_t(\z)$ and the target distribution as $\pi(\z)$. The trajectory of \ac{ld} in \cref{equ:ld} is typically specified by its transition kernel ${\mK}(\z|\z')$. The process starts with drawing $\z_0$ from $\pi_0(\z)$ and iteratively sample $\z_{t}$ at the $t$-th iteration from the transition kernel conditioned on $\z_{t-1}$, \ie, $\pi_{t} (\z) = {\mK}\pi_{t - 1}(\z)$, where ${\mK}\pi_{t - 1}(\z) := \int {\mK}(\z | \z') \pi_{t - 1}(\z') d\z'$. Recursively, $\pi_t = {\mK}_t \pi_0$, where ${\mK}_t$ denotes the $t$-step transition kernel.
\ac{ld} can therefore be viewed as approximating a fixed point update in a non-parametric fashion since the target distribution $\pi$ is a stationary distribution $\pi(\z) := \int {\mK}(\z | \z') \pi(\z') d\z', \forall \z$. This motivates several works for more general approximations of this update \cite{li2017approximate, xie2018cooperative, ruiz2019contrastive} with the help of neural samplers. 

Inspired by \cite{li2017approximate}, we propose to use the following framework for amortizing the \ac{ld} in learning the \ac{lebm}. Formally, let ${\mQ} = \{q_{\bP}\}$ be the set of amortized samplers parameterized by ${\bP}$. Given the transition kernel $\mK$, the goal is to find a sampler $q_{{\bP}^*}$ to closely approximate the target distribution $\pi$. This can be achieved by iteratively distill $T$-step transitions of \ac{ld} into $q_{\bP}$:
\begin{equation}
\label{equ:qphi}
q_{{\bP}_{k}} \gets 
 \argmin_{q_{\bP} \in {\mQ}} {\mD}[q_{{\bP}_{k-1}, T}||q_{\bP}],~
q_{{\bP}_{k-1}, T} := {\mK}_T q_{{\bP}_{k-1}},~
q_{\bP_0} \approx \pi_0,~k=0,...,K-1. 
\end{equation}
where ${\mD}[\cdot || \cdot]$ is the \ac{kld} measure between distributions. Concretely, \cref{equ:qphi} means that to recover the target distribution $\pi$, instead of using long-run \ac{ld}, we can repeat the following steps: i) employ a $T$-step short-run \ac{ld} initialized with the current sampler $q_{{\bP}_{k-1}}$ to approximate ${\mK}_T q_{{\bP}_{k-1}}$ as the target distribution of the current sampler, and ii) update the current sampler $q_{{\bP}_{k-1}}$ to $q_{{\bP}_{k}}$. The correct convergence of $q_{\bP}$ to $\pi$ with \cref{equ:qphi} is supported by the standard theory of Markov chains \cite{cover1999elements}, which suggests that the update in \cref{equ:qphi} is monotonically decreasing in terms of \ac{kld}, 
${\mD} [q_{{\bP}_k} || \pi ] \leq {\mD} [q_{{\bP}_{k-1}} || \pi ]$. We refer to \cref{app:mdkld} for a detailed explanation and discussion of this statement. In practice, one can apply gradient-based methods to minimize ${\mD}[q_{{\bP}_{k-1}, T}||q_{\bP}]$ and approximate \cref{equ:qphi} for the update from $q_{{\bP}_{k-1}}$ to $q_{{\bP}_k}$. The above formulation provides a generic and flexible framework for amortizing the potentially long MCMC. 

\paragraph{Diffusion-based amortization}
To avoid clutter, we simply write $q_{{\bP}_{k-1}, T}$ as $q_T$. We can see that
\begin{equation}
\label{equ:cent}
\begin{aligned}
\argmin_{q_{\bP}} {\mD}[q_T||q_{\bP}] &=
\argmin_{q_{\bP}} -{\mH}(q_{T}) + {\mH}(q_{T}, q_{\bP}) = 
\argmin_{q_{\bP}} -\E_{q_{T}}\left[
\log q_{\bP}
\right],
\end{aligned}
\end{equation}
where ${\mH}$ represents the entropy of distributions. The selection of the sampler $q_{\bP}$ is a matter of design. According to \cref{equ:cent}, we may expect the following properties from $q_{\bP}$: i) having analytically tractable expression of the exact value or lower bound of log-likelihood, ii) easy to draw samples from and iii) capable of close approximation to the given distribution $\{q_T\}$. In practice, iii) is important for the convergence of \cref{equ:qphi}. If $q_{\bP}$ is far away from $q_T$ in each iteration, then non-increasing \ac{kld} property ${\mD} [q_{{\bP}_k} || \pi ] \leq {\mD} [q_{{\bP}_{k-1}} || \pi ]$ might not hold, and the resulting amortized sampler would not converge to the true target distribution $\pi(\z)$. 

For the choice of $q_{\bP}$, let us consider distilling the gradient field of $q_T$ in each iteration, so that the resulting sampler is close to the $q_T$ distribution. This naturally points to the \acp{ddpm} \cite{ho2020denoising}. To be specific, learning a \ac{ddpm} with ${\be}$-prediction parameterization is equivalent to fitting the finite-time marginal of a sampling chain resembling annealed Langevin dynamics \cite{song2019generative,ho2020denoising,song2020improved}. Moreover, it also fulfills i) and ii) of the desired properties mentioned above. We can plug in the objective of \ac{ddpm} (\cref{sec:ddpm}), which is a lower bound of $\log q_{\bP}$, to obtain the gradient-based update rule for $q_{\bP}$:
\begin{equation}
\label{equ:q_upd}
\begin{aligned}
{\bP}_{k-1}^{(i+1)} \gets {\bP}_{k-1}^{(i)} - \eta \nabla_{\bP} \E_{{\be},\lambda}\left[\|{\be}(\z_\lambda) - {\be} \|_2^2\right],~
{\bP}_{k}^{(0)} \gets {\bP}_{k-1}^{(M)},~i=0,1,...,M-1
\end{aligned}
\end{equation}
where ${\be} \sim \N(0, \I_d)$. $\lambda$ is drawn from a distribution of log noise-to-signal ratio $p(\lambda)$. $\eta$ is the step size for the update, and $M$ is the number of iterations needed in \cref{equ:q_upd}. In practice, we find that when amortizing the \ac{ld} sampling chain, a light-weight \ac{ddpm} $q_{\bP}$ updated with very small $M$, \ie, few steps of \cref{equ:q_upd} iteration, approximates \cref{equ:qphi} well. We provide a possible explanation using the Fisher information by scoping the asymptotic behavior of this update rule in the \cref{app:diff_amor}. We term the resulting sampler as \ac{damc}.

\subsection{Approximate \texorpdfstring{\acs{mle}}{} with \texorpdfstring{\acs{damc}}{}}
\label{sec:app_mle}
In this section, we show how to integrate the \ac{damc} sampler into the learning framework of \ac{lebm} and form a symbiosis between these models. Given a set of $N$ training samples $\{\x_i\}_{i=1}^N$ independently drawn from the unknown data distribution $p_{\text{data}}(\x)$, the model $p_{\bT}$ (\cref{sec:lebm}) can be trained by maximizing the log-likelihood over training samples ${\mL}({\bT})=\frac{1}{N} \sum_{i=1}^{N} \log p_{\bT}\left(\x_{i}\right)$. Doing so typically requires computing the gradients of ${\mL}({\bT})$, where for each $\x_i$ the learning gradient satisfies:
\begin{equation}
\label{equ:mle}
\begin{aligned}
\nabla_{\bT} \log p_{\bT}(\x_i) &= \E_{p_{\bT}(\z_i|\x_i)}\left[\nabla_{\bT} \log p_{\bT}(\z_i, \x_i)\right] \\
&=(
\underbrace{
\E_{p_{\bT}(\z_i|\x_i)} 
    \left[\nabla_{\bA} f_{\bA}(\z_i) \right] - 
\E_{p_{\bA}(\z_i)} 
    \left[\nabla_{\bA} f_{\bA}(\z_i) \right]
}_{\delta_{\bA} (\x_i)}, 
\underbrace{
\E_{p_{\bT}(\z_i|\x_i)} 
    \left[\nabla_{\bB} \log p_{\bB}(\x_i | \z_i) \right] 
}_{\delta_{\bB} (\x_i)}). 
\end{aligned}
\end{equation}

Intuitively, based on the discussion in \cref{sec:lebm} and \cref{sec:damc}, we can approximate the distributions in \cref{equ:mle} by drawing samples from
$
\left[\z_i | \x_i\right] \sim {\mK}_{T, \z_i | \x_i} q_{{\bP}_{k}}(\z_i | \x_i),~
\z_i \sim {\mK}_{T, \z_i} q_{{\bP}_{k}}(\z_i),
$
to estimate the expectations and hence the learning gradient. Here we learn the \ac{damc} samplers $q_{{\bP}_{k}}(\z_i | \x_i)$ and $q_{{\bP}_{k}}(\z_i)$ for the posterior and prior sampling chain, respectively. ${\bP}_{k}$ represents the current sampler as in \cref{sec:damc}; ${\mK}_{T, \z_i | \x_i}$ and ${\mK}_{T, \z_i}$ are the transition kernels for posterior and prior sampling chain. Equivalently, this means to better estimate the learning gradients we can i) first draw approximate posterior and prior MCMC samples from the current $q_{{\bP}_k}$ model, and ii) update the approximation of the prior $p_{\bA}(\z)$ and posterior $p_{\bT}(\z | \x)$ distributions with additional $T$-step \ac{ld} initialized with $q_{{\bP}_k}$ samples. These updated samples are closer to $p_{\bT}(\z_i|\x_i)$ and $p_{\bA}(\z_i)$ compared with short-run \ac{ld} samples based on our discussion in \cref{sec:ddpm}. Consequently, the diffusion-amortized \ac{ld} samples provide a generally better estimation of the learning gradients and lead to better performance, as we will show empirically in \cref{sec:exprs}. After updating ${\bT}=({\bA},{\bB})$ based on these approximate samples with \cref{equ:mle}, we can update $q_{{\bP}_k}$ with \cref{equ:q_upd} to distill the sampling chain into $q_{{\bP}_{k+1}}$. As shown in \cref{fig:method}, we can see that the whole learning procedure iterates between the approximate \ac{mle} of $p_{\bT}$ and the amortization of MCMC with $q_{\bP}$. We refer to \cref{app:disc_lalg} for an extended discussion of this procedure.

After learning the models, we can use either \ac{damc} or \ac{lebm} for prior sampling. For \ac{damc}, we may draw samples from $q_{\bP}(\z_i)$ with \cref{equ:ac_sampler}. Prior sampling with \ac{lebm} still requires short-run \ac{ld} initialized from $\N(0, \I_d)$. For posterior sampling, we may sample from ${\mK}_{T, \z_i | \x_i} q_{{\bP}_{k}}(\z_i | \x_i)$, \ie, first draw samples from \ac{damc} and then run few steps of \ac{ld} to obtain posterior samples.

\begin{algorithm}[!htbp]
\caption{\textbf{Learning algorithm of \ac{damc}.}}
\label{alg:learn_algo}
\KwIn{
initial parameters $({\bA}, {\bB}, {\bP})$; 
learning rate $\eta = (\eta_{\bA}, \eta_{\bB}, \eta_{\bP})$;
observed examples $\{\x^{(i)}\}_{i=1}^N$;
prob. of uncond. training $p_{\rm uncond}$ for the \ac{damc} sampler.}
\KwOut{
$\left({\bT}^{(K)} = \{{\bA}^{(K)}, {\bB}^{(K)}\}, {\bP}^{(K)}\right)$.
}
\BlankLine	
\For{$k=0:K-1$}{
Sample a minibatch of data $\{\x^{(i)}\}_{i=1}^B$;

\textbf{Draw \ac{damc} samples:} For each $\x^{(i)}$, draw $\z_{+}^{(i)}$ and $\z_{-}^{(i)}$ from $q_{{\bP}_{k}}(\z_i | \x_i)$. 

\textbf{Prior \ac{ld} update:} For each $\x^{(i)}$, update $\z_{-}^{(i)}$ using \cref{equ:ld}, with $\pi(\z_i) = p_{{\bA}^{(k)}}(\z_i)$;
	    
\textbf{Posterior \ac{ld} update:} For each $\x^{(i)}$, update $\z_{+}^{(i)}$ using \cref{equ:ld}, with $\pi(\z_i) = p_{{\bB}^{(k)}}(\z_i|\x_i)$;

\textbf{Update ${\bT}^{(k)}$:} Update ${\bA}^{(k)}$ and ${\bB}^{(k)}$ using Monte-Carlo estimates (\ie, Monte-Carlo average) of \cref{equ:mle} with $\{\z_{+}^{(i)}\}_{i=1}^B$ and $\{\z_{-}^{(i)}\}_{i=1}^B$.

\textbf{Update ${\bP}^{(k)}$:} Update ${\bP}^{(k)}$ using \cref{equ:q_upd} with $p_{\rm uncond}$ and $\{\z_{+}^{(i)}\}_{i=1}^B$ as the target.
}
\end{algorithm}

\subsection{Implementation}
\label{sec:implem}
In order to efficiently model both $q_{{\bP}_{k}}(\z_i | \x_i)$ and $q_{{\bP}_{k}}(\z_i)$, we follow the method of \cite{ho2021classifier} to train a single neural network to parameterize both models, where $q_{{\bP}_{k}}(\z_i | \x_i)$ can be viewed as a conditional \ac{ddpm} with the embedding of $\x_i$ produced by an encoder network as its condition, and $q_{{\bP}_{k}}(\z_i)$ an unconditional one. For $q_{{\bP}_{k}}(\z_i)$, we can simply input a null token $\emptyset$ as its condition when predicting the noise $\be$. We jointly train both models by randomly nullifying the inputs with the probability $p_{\rm uncond}=0.2$. During training, we use samples from $q_{{\bP}_{k}}(\z_i | \x_i)$ to initialize both prior and posterior updates. For the posterior and prior \ac{damc} samplers, we set the number of diffusion steps to $100$. The number of iterations in \cref{equ:q_upd} is set to $M=6$ throughout the experiments. The \ac{ld} runs $T=30$ and $T=60$ iterations for posterior and prior updates during training with a step size of $s=0.1$. For test time sampling from ${\mK}_{T, \z_i | \x_i} q_{{\bP}_{k}}(\z_i | \x_i)$, $T=10$ for the additional \ac{ld}. For a fair comparison, we use the same \ac{lebm} and generator as in \cite{pang2020learning,xiao2022adaptive} for all the experiments. We summarize the learning algorithm in \cref{alg:learn_algo}. Please see  \cref{app:model_and_arch,app:pcode} for network architecture and further training details, as well as the pytorch-style pseudocode of the algorithm.

\section{Experiments}
\label{sec:exprs}
In this section, we are interested in the following questions: (i) How does the proposed method compare with its previous counterparts (\eg, purely MCMC-based or variational methods)? (ii) How is the scalability of this method? (iii) How are the time and parameter efficiencies? (iv) Does the proposed method provide a desirable latent space? To answer these questions, we present a series of experiments on benchmark datasets including MNIST \cite{lecun1998mnist}, SVHN \cite{netzer2011reading}, CelebA64 \cite{liu2015deep}, CIFAR-10 \cite{krizhevsky2009learning}, CelebAMask-HQ \cite{CelebAMask-HQ}, FFHQ \cite{karras2019style} and LSUN-Tower \cite{yu2015lsun}. As to be shown, the proposed method demonstrates consistently better performance in various experimental settings compared with previous methods. We refer to \cref{app:data_n_expr} for details about the experiments.

\begin{table}[!t]
\caption{\textbf{MSE($\downarrow$) and FID($\downarrow$) obtained from models trained on different datasets}. The FID scores are computed based on 50k generated images and training images for the first three datasets and 5k images for the CelebA-HQ dataset. The MSEs are computed based on unseen testing images. We highlight our model results in {\color{tgray} gray} color. The best and second-best performances are marked in bold numbers and underlines, respectively; tables henceforth follow this format. *\cite{xiao2022adaptive} uses a prior model with $4$x parameters compared with \cite{pang2020learning} and ours.}
\label{table:fid_recon}
\centering
\begin{tabular}{ccccccccc}
    \toprule
       \multirow{2.5}{*}{Model}
     & \multicolumn{2}{c}{SVHN} 
     & \multicolumn{2}{c}{CelebA}   
     & \multicolumn{2}{c}{CIFAR-10}
     & \multicolumn{2}{c}{CelebA-HQ}\\
     \cmidrule(r){2-3} \cmidrule(r){4-5} 
     \cmidrule(r){6-7} \cmidrule(r){8-9}
{} & MSE & FID & MSE & FID & MSE & FID & MSE & FID\\
\midrule
VAE \cite{kingma2013auto} &0.019 &46.78&0.021&65.75&0.057&106.37
&0.031 &180.49\\
2s-VAE \cite{dai2019diagnosing} &0.019&42.81&0.021&44.40&0.056&72.90
&-&-\\
RAE \cite{ghosh2019variational}& 0.014&40.02&0.018&40.95&0.027&74.16
&-&-\\
NCP-VAE \cite{aneja2021contrastive} &0.020&33.23&0.021&42.07&0.054&78.06
&-&-\\
\midrule
Adaptive ${\rm CE}^{*}$ \cite{xiao2022adaptive} &\underline{0.004}&26.19&\underline{0.009}&35.38&\textbf{0.008}&65.01&-&-\\
\midrule
ABP \cite{han2017alternating}
&-&49.71&-&51.50&0.018&90.30&0.025&160.21\\
SRI \cite{nijkamp2019learning} 
&0.018&44.86&0.020&61.03&-&-&-&-\\
SRI (L=5) \citep{nijkamp2019learning} 
&0.011&35.32&0.015&47.95&-&-&-&-\\
LEBM \cite{pang2020learning} 
&0.008&29.44&0.013&37.87&0.020&70.15&\underline{0.025}&133.07\\
\midrule
\rowcolor{gray}
Ours-\ac{lebm}&
{} & \underline{21.17}&
{} & \underline{35.67}&
{} & \underline{60.89}&
{} &\underline{89.54}\\
\rowcolor{gray}
Ours-\ac{damc}&
\multirow{-2.}{*}{\textbf{0.002}} & \textbf{18.76}&
\multirow{-2.}{*}{\textbf{0.005}} & \textbf{30.83}&
\multirow{-2.}{*}{\underline{0.015}} & \textbf{57.72}&
\multirow{-2.}{*}{\textbf{0.023}}&\textbf{85.88}\\
\bottomrule
\end{tabular}
\vspace{-10pt}
\end{table}

\subsection{Generation and Inference: Prior and Posterior Sampling}
\paragraph{Generation and reconstruction} We evaluate the quality of the generated and reconstructed images to examine the sampling quality of \ac{damc}. Specifically, we would like to check i) how well does \ac{damc} fit the seen data, ii) does \ac{damc} provide better MCMC samples for learning \ac{lebm} and iii) the generalizability of \ac{damc} on unseen data. We check the goodness of fit of \ac{damc} by evaluating the quality of the images generated with \ac{damc} prior samples. If \ac{damc} does provide better MCMC samples for learning \ac{lebm}, we would expect better fitting of data and hence an improved generation quality of \ac{lebm}. We evaluate the performance of posterior sampling given unseen testing images by examining the reconstruction error on testing data. We benchmark our model against a variety of previous methods in two groups. The first group covers competing methods that adopt the variational learning scheme, including VAE \citep{kingma2013auto}, as well as recent two-stage methods such as 2-stage VAE \citep{dai2019diagnosing}, RAE \citep{ghosh2019variational} and NCP-VAE \citep{aneja2021contrastive}, whose prior distributions are learned with posterior samples in a second stage after the generator is trained. The second group includes methods that adopt MCMC-based sampling. It includes \ac{abp} \citep{han2017alternating}, \ac{sri} from \citep{nijkamp2019learning} and the vanilla learning method of \ac{lebm} \citep{pang2020learning}, which relies on short-run \ac{ld} for both posterior and prior sampling. We also compare our method with the recently proposed Adaptive CE \cite{xiao2022adaptive}. It learns a series of \acp{lebm} adaptively during training, while these \acp{lebm} are sequentially updated by density ratio estimation instead of \ac{mle}. To make fair comparisons, we follow the same evaluation protocol as in \citep{pang2020learning,xiao2022adaptive}.

\begin{figure}
    \centering
    \subfigure[SVHN]{
        \includegraphics[width=0.23\textwidth]{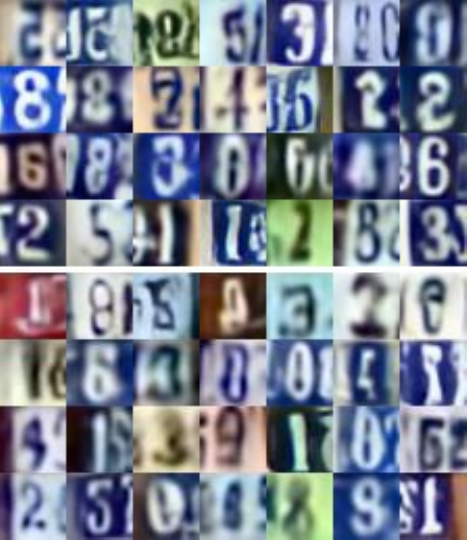}
    }
    \subfigure[CelebA]{
        \includegraphics[width=0.23\textwidth]{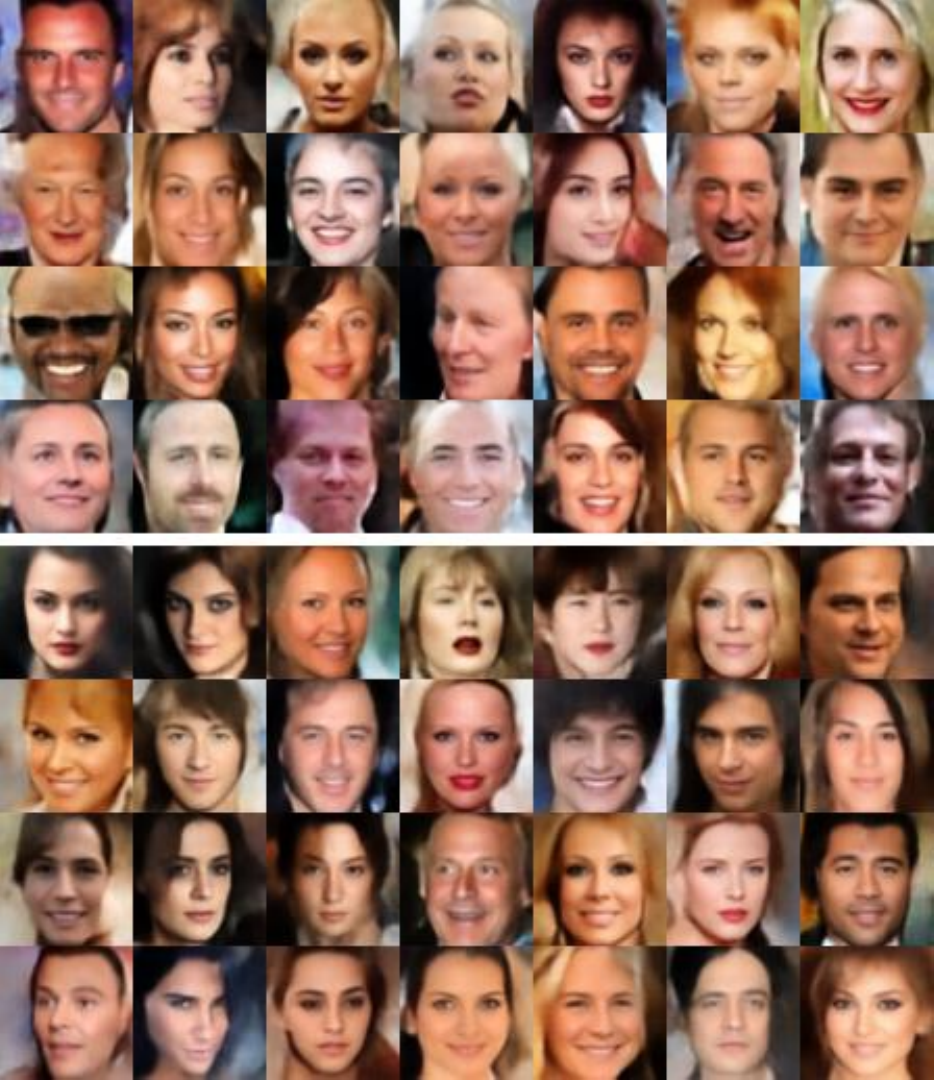}
    }
    \subfigure[CIFAR-10]{
        \includegraphics[width=0.23\textwidth]{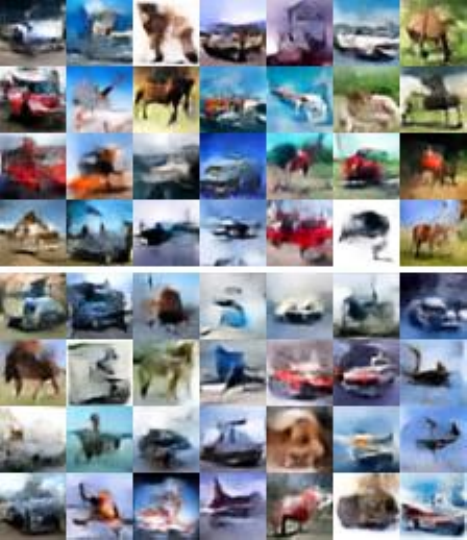}
    }
    \subfigure[CelebA-HQ]{
        \includegraphics[width=0.23\textwidth]{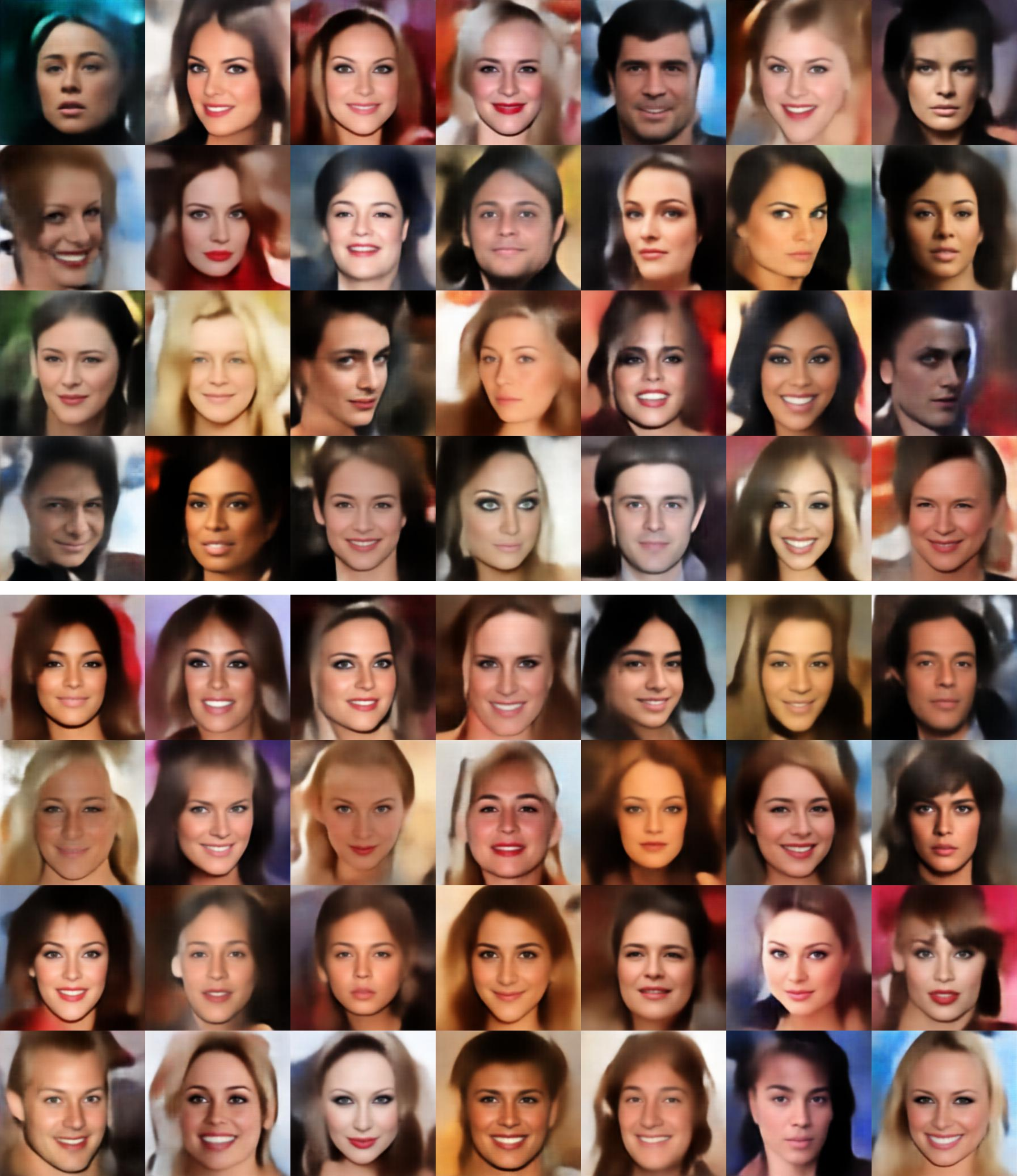}
    }
    \caption{\textbf{Samples generated from the \ac{damc} sampler and \ac{lebm}} trained on SVHN, CelebA, CIFAR-10 and CelebA-HQ datasets. In each sub-figure, the first four rows are generated by the \ac{damc} sampler. The last four rows are generated by \ac{lebm} trained with the \ac{damc} sampler.}
    \label{fig:gen}
    \vspace{-10pt}
\end{figure}

For generation, we report the FID scores \cite{heusel2017gans} in \cref{table:fid_recon}. We observe that i) the \ac{damc} sampler, denoted as Ours-\ac{damc}, provides superior generation performance compared to baseline models, and ii) the \ac{lebm} learned with samples from \ac{damc}, denoted as Ours-\ac{lebm}, demonstrates significant performance improvement compared with the \ac{lebm} trained with short-run \ac{ld}, denoted as \ac{lebm}. These results confirm that \ac{damc} is a reliable sampler and indeed partly addresses the learning issue of \ac{lebm} caused by short-run \ac{ld}. We would like to point out that the improvement is clearer on the CelebAMask-HQ dataset, where the input data is of higher dimension ($256\times256$) and contains richer details compared with other datasets. This illustrates the superiority of \ac{damc} sampler over short-run \ac{ld} when the target distribution is potentially highly multi-modal. We show qualitative results of generated samples in \cref{fig:gen}, where we observe that our method can generate diverse, sharp and high-quality samples. For reconstruction, we compare our method with baseline methods in terms of MSE in \cref{table:fid_recon}. We observe that our method demonstrates competitive reconstruction error, if not better, than competing methods do. Additional qualitative results of generation and reconstruction are presented in \cref{app:gen_app,app:recon_app}.

\begin{wraptable}{r}{0.5\linewidth}
\vspace{-15pt}
\centering
\resizebox{\linewidth}{!}{
\begin{tabular}{lcccc}
\toprule
  \multirow{2.5}{*}{Model}
& \multicolumn{2}{c}{FFHQ} 
& \multicolumn{2}{c}{LSUN-T} \\
\cmidrule(r){2-3} \cmidrule(r){4-5}
{} & MSE & FID & MSE & FID \\
\midrule
Opt. \cite{abdal2019image2stylegan} & 0.055 & 149.39 & 0.080 & 240.11 \\
Enc. \cite{zhu2016generative} & \underline{0.028} & \underline{62.32} & 0.079 & 132.41 \\
\midrule
\cite{pang2020learning} w/ 1x & 0.054 & 149.21 & 0.072 & 239.51 \\
\cite{pang2020learning} w/ 2x & 0.039 & 101.59 & 0.066 & 163.20 \\
\cite{pang2020learning} w/ 4x & 0.032 & 84.64 & \underline{0.059} & \underline{111.53} \\
\midrule
\rowcolor{gray}
Ours & \textbf{0.025} & \textbf{52.85} 
     & \textbf{0.059} & \textbf{80.42} \\
\bottomrule
\end{tabular}
}
\caption{\textbf{MSE($\downarrow$) and FID($\downarrow$) for GAN inversion on different datasets}. Opt. and Enc. denotes the optimization-based and encoder-based methods.}
\label{tab:gan_inv}
\vspace{-10pt}
\end{wraptable}

\begin{figure}[!t]
    \centering
    \subfigure[Observation]{
        \includegraphics[width=0.95\textwidth]{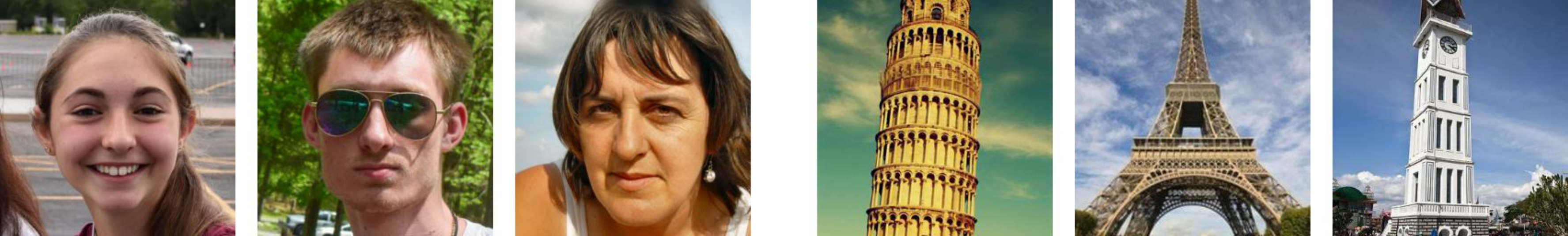}
    }
    \quad
    \subfigure[Inversion]{
        \includegraphics[width=0.95\textwidth]{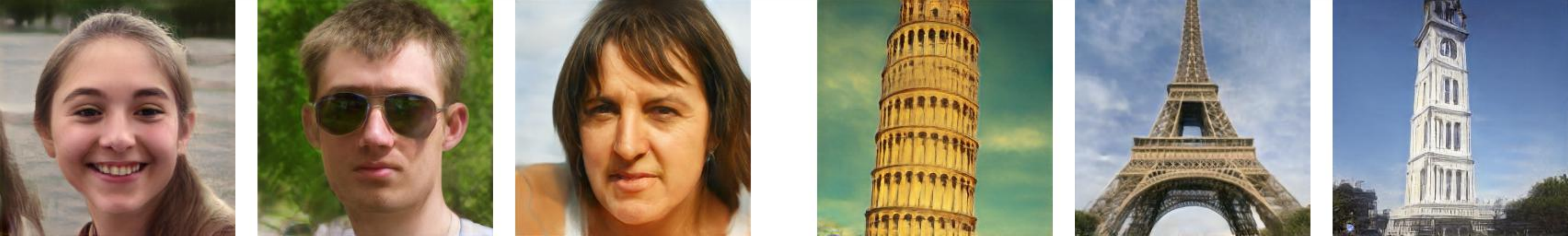}
    }
    \caption{\textbf{Qualitative results of StyleGAN inversion using the \ac{damc} sampler.} In each sub-figure, the left panel contain samples from the FFHQ dataset, and the right panel contains samples from the LSUN-T dataset.}
    \label{fig:inv}
    \vspace{-10pt}
\end{figure}

\paragraph{GAN inversion} We have examined the scalability of our method on the CelebAMask-HQ dataset. Next, we provide more results on high-dimensional and highly multi-modal data by performing GAN inversion \cite{xia2022gan} using the proposed method. Indeed, we may regard GAN inversion as an inference problem and a special case of posterior sampling. As a suitable testbed, the StyleGAN structure \cite{karras2019style} is specifically considered as our generator in the experiments: \cite{abdal2019image2stylegan} points out that to effectively infer the latent representation of a given image, the GAN inversion method needs to consider an extended latent space of StyleGAN, consisting of 14 different 512-dimensional latent vectors. 
We attempt to use the \ac{damc} sampler for GAN inversion. We benchmark our method against i) learning an encoder that maps a given image
to the latent space \cite{zhu2016generative}, which relates to the variational methods for posterior inference, ii) optimizing a random initial latent code by minimizing the reconstruction error and perceptual loss \cite{abdal2019image2stylegan}, which can be viewed as a variant of \ac{ld} sampling, and iii) optimizing the latent code by minimizing both the objectives used in ii) and the energy score provided by \ac{lebm}. We use the pretrained weights provided by \cite{zhu2020domain} for the experiments. Both the \ac{damc} sampler and the encoder-based method are augmented with $100$ post-processing optimization iterations. We refer to \cref{app:data_n_expr} for more experiment details. We test \ac{lebm}-based inversion with different optimization iterations. To be specific, 1x, 2x, and 4x represent 100, 200, and 400 iterations respectively. We can see in \cref{tab:gan_inv} that \ac{damc} performs better than all the baseline methods on the unseen testing data, which supports the efficacy of our method in high-dimensional settings. We provide qualitative results in \cref{fig:inv}.

\paragraph{Parameter efficiency and sampling time}
One potential disadvantage of our method is its parameter inefficiency for introducing an extra \ac{ddpm}. Fortunately, our models are in the latent space so the network is lightweight. To be specific, on CIFAR-10 dataset the number of parameters in the \ac{ddpm} is only around $10\%$ (excluding the encoder) of those in the generator. The method has competitive time efficiency. With the batch size of $64$, the \ac{damc} prior sampling takes $0.3s$, while $100$ steps of short-run \ac{ld} with \ac{lebm} takes $0.2s$. The \ac{damc} posterior sampling takes $1.0s$, while \ac{lebm} takes $8.0s$. Further discussions about the limitations can be found in the \cref{app:limit}.  

\subsection{Analysis of Latent Space}

\begin{figure}
    \centering
    \subfigure[200 steps]{
        \includegraphics[width=0.47\textwidth]{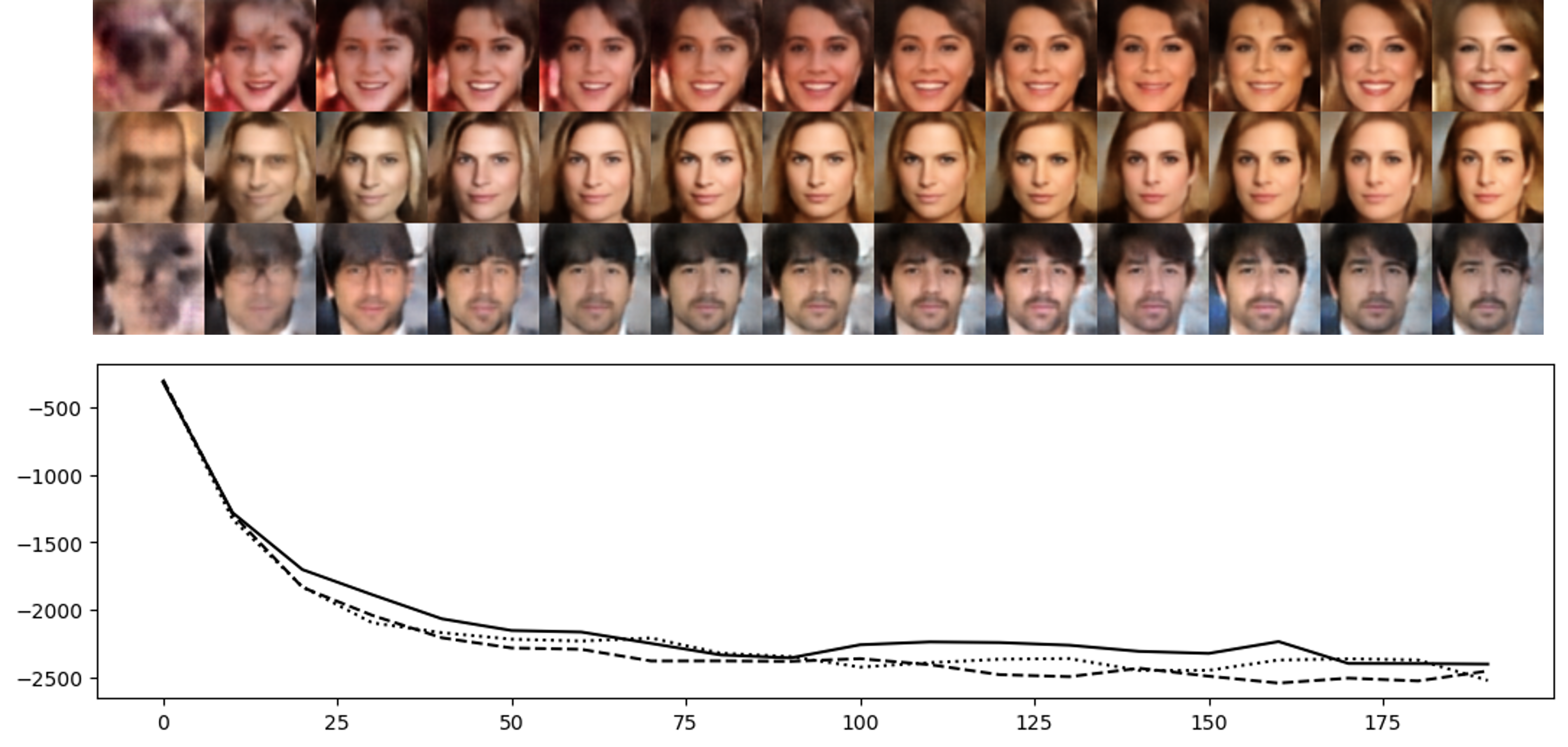}
    }
    \subfigure[2500 steps]{
        \includegraphics[width=0.47\textwidth]{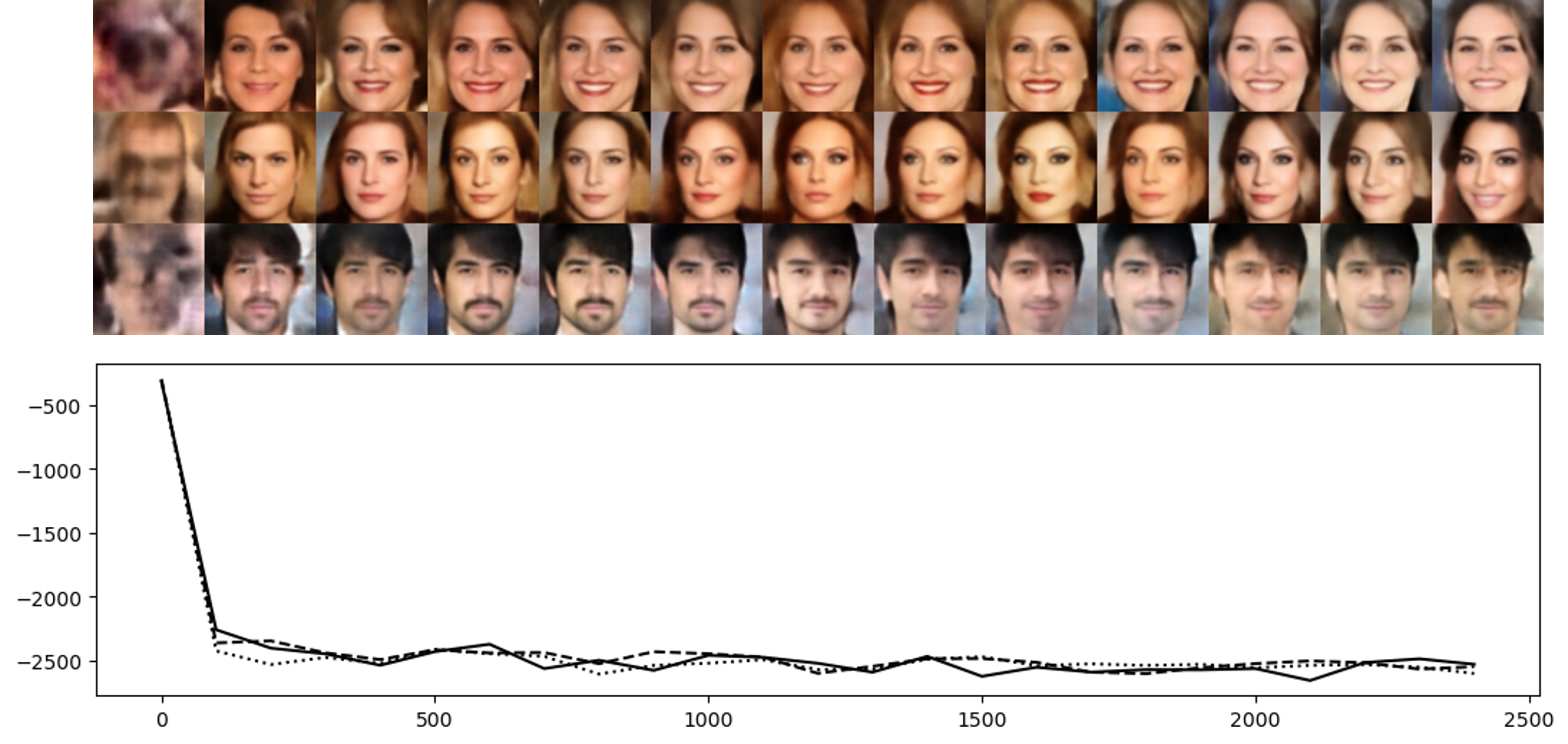}
    }
    \caption{\textbf{Transition of Markov chains initialized from $\N(0, \I_d)$ towards $p_{\bA}(\z)$.} We present results by running \ac{ld} for 200 and 2500 steps. In each sub-figure, the top panel displays the trajectory in the data space uniformly sampled along the chain. The bottom panel shows the energy score $f_{\bA}(\z)$ over the iterations.}
    \label{fig:anl}
    \vspace{-10pt}
\end{figure}

\paragraph{Long-run langevin transition} 
In this section, we examine the energy landscape induced by the learned \ac{lebm}. We expect that a well-trained $p_{\bA}(\z)$ fueled by better prior and posterior samples from the \ac{damc} sampler would lead to energy landscape with regular geometry.
In \cref{fig:anl}, we visualize the transition of \ac{ld} initialized from $\N(0, \I_d)$ towards $p_{\bA}(\z)$ on the model trained on the CelebA dataset. Additional visualization of transitions on SVHN and CIFAR-10 datasets can be found in the \cref{app:trans}. The \ac{ld} iterates for $200$ and $2500$ steps, which is longer than the \ac{ld} within each training iteration ($60$ steps). For the $200$-step set-up, we can see that the generation quality quickly improves by exploring the local modes (demonstrating different facial features, \eg, hairstyle, facial expression and lighting). For the $2500$-step long-run set-up, we can see that the \ac{ld} produces consistently valid results without the oversaturating issue of the long-run chain samples \cite{nijkamp2020anatomy}. These observations provide empirical evidence that the \ac{lebm} is well-trained.

\paragraph{Anomaly detection}
We further evaluate how the \ac{lebm} learned by our method could benefit the anomaly detection task. With properly learned models, the posterior $p_{{\bT},{\bP}}(\z|\x)$ could form a discriminative latent space that has separated probability densities for in-distribution (normal) and out-of-distribution (anomalous) data. Given the testing sample $\x$, we use un-normalized log joint density $p_{{\bT},{\bP}}(\z | \x) \propto p_{{\bT},{\bP}}(\x, \z) \approx p_{\bB}(\x | \z) p_{\bA}(\z)|_{\z \sim {\mK}_{T, \z|\x} q_{\bP}(\z|\x)}$ as our decision function. 
This means that we draw samples from ${\mK}_{T, \z|\x} q_{\bP}(\z|\x)$ and compare the corresponding reconstruction errors and energy scores. 
A higher value of log joint density indicates a higher probability of the test sample being a normal sample. To make fair comparisons, we follow the experimental settings in \cite{pang2020learning,xiao2022adaptive,kumar2019maximum,zenati2018efficient} and train our model on MNIST with one class held out as an anomalous class. We consider the baseline models that employ MCMC-based or variational inferential mechanisms.
\cref{table:auprc} shows the results of
AUPRC scores averaged over the last 10 trials. We observe significant improvements in our method over the previous counterparts.

\begin{table}
\caption{\textbf{AUPRC($\uparrow$) scores for unsupervised anomaly detection on MNIST}. Numbers are taken from \citep{xiao2022adaptive}. Results of our model are averaged over the last 10 trials to account for variance.}
\label{table:auprc}
\centering
\resizebox{\linewidth}{!}{
\begin{tabular}{cccccc}
\toprule
Heldout Digit & 1 & 4 & 5 & 7 & 9 \\
\midrule
VAE \cite{kingma2013auto} 
&0.063 &0.337 &0.325 &0.148 &0.104\\
ABP \cite{han2017alternating}
&$0.095\pm 0.03$ & $0.138\pm 0.04$
&$0.147\pm 0.03$ & $0.138\pm 0.02$ & $0.102\pm 0.03$ \\
MEG \cite{kumar2019maximum} 
&$0.281\pm 0.04$ & $0.401\pm 0.06$ 
&$0.402\pm 0.06$ & $0.290\pm 0.04$ & $0.342\pm 0.03$ \\
BiGAN-$\sigma$ \cite{zenati2018efficient} 
&$0.287\pm 0.02$ & $0.443\pm 0.03$ 
&$0.514\pm 0.03$ & $0.347\pm 0.02$ & $0.307\pm 0.03$ \\
LEBM \cite{pang2020learning} 
&$0.336\pm 0.01$ & $0.630\pm 0.02$
&$0.619\pm 0.01$ & $0.463\pm 0.01$ & $0.413\pm 0.01$ \\
Adaptive CE \cite{xiao2022adaptive} 
&\underline{$0.531\pm 0.02$} 
&\underline{$0.729\pm 0.02$}
&\underline{$0.742\pm 0.01$} 
&\underline{$0.620\pm 0.02$} 
&\underline{$0.499\pm 0.01$}\\
\midrule
\rowcolor{gray}
Ours
&$\textbf{0.684} \pm \textbf{0.02}$ 
&$\textbf{0.911} \pm \textbf{0.01}$ 
&$\textbf{0.939} \pm \textbf{0.02}$ 
&$\textbf{0.801} \pm \textbf{0.01}$ 
&$\textbf{0.705} \pm \textbf{0.01}$ \\
\bottomrule
\end{tabular}
}
\vspace{-10pt}
\end{table}

\subsection{Ablation Study}
In this section, we conduct ablation study on several variants of the proposed method. Specifically, we would like to know: i) what is the difference between the proposed method and directly training a \ac{ddpm} in a fixed latent space? ii) What is the role of \ac{lebm} in this learning scheme? iii) Does \ac{damc} effectively amortize the sampling chain? We use CIFAR-10 dataset for the ablative experiments to empirically answer these questions. More ablation studies can be found in \cref{app:quati_res}.

\paragraph{Non-Amortized \ac{ddpm} vs. \ac{damc}} We term directly training a \ac{ddpm} in a fixed latent space as the non-amortized \ac{ddpm}. To analyze the difference between non-amortized \ac{ddpm} and \ac{damc}, we first train a \ac{lebm} model with persistent long-run chain sampling \cite{tieleman2008training} and use the trained model to obtain persistent samples for learning the non-amortized \ac{ddpm}. In short, the non-amortized \ac{ddpm} can be viewed as directly distilling the long-run \ac{mcmc} sampling process, instead of progressively amortizing the chain. We present the FID and MSE of the non-amortized model (NALR) in \cref{table:abl}.
We observe that directly amortizing the long-run chain leads to degraded performance compared with the proposed method. The results are consistently worse for both posterior and prior sampling and the learned \acp{lebm}, which verify the effectiveness of the proposed iterative learning scheme.

\paragraph{The contribution of \ac{lebm}} One may argue that since we have the \ac{damc} as a powerful sampler, it might not be necessary to jointly learn \ac{lebm} in the latent space. To demonstrate the necessity of this joint learning scheme, we train a variant of \ac{damc} by replacing the \ac{lebm} with a Gaussian prior. The results are presented in \cref{table:abl}. We observe that models trained with non-informative Gaussian prior obtain significantly worse generation results. It suggests that \ac{lebm} involved in the learning iteration provides positive feedback to the \ac{damc} sampler. Therefore, we believe that it is crucial to jointly learn the \ac{damc} and \ac{lebm}. 

\paragraph{Vanilla sampling vs. \ac{damc}} We compare the vanilla sampling process of each model with \ac{damc}. The vanilla sampling typically refers to short-run or long-run \ac{ld} initialized with $\N(0, \I_d)$. We also provide results of learning \ac{lebm} using variational methods for comparison. We can see in \cref{table:abl} that sampling with \ac{damc} shows significantly better scores of the listed models, compared with vanilla sampling. The result is even better than that of the persistent chain sampler (V. of NALR-LEBM). This indicates that \ac{damc} effectively amortizes the sampling chain.
Comparing \ac{damc} sampler with the variational sampler also indicates that \ac{damc} is different from general variational approximation: it benefits from its connection with \ac{ld} and shows better expressive power.

\begin{table}
\caption{\textbf{Ablation study on CIFAR-10 dataset}. VI denotes learning \ac{lebm} using variational methods. SR denotes learning \ac{lebm} with short-run \ac{ld}. DAMC-G replaces the \ac{lebm} in DAMC-LEBM with a standard Gaussian distribution. NALR denotes the non-amortized \ac{ddpm} setting. For each set-up, we provide results using the vanilla sampling method, denoted as V, and the ones using the \ac{damc} sampler, denoted as D.}
\label{table:abl}
\centering
\begin{tabular}{ccccccccccc}
\toprule
  \multirow{2.5}{*}{Model}
& \multicolumn{2}{c}{VI-LEBM} 
& \multicolumn{2}{c}{SR-LEBM}   
& \multicolumn{2}{c}{DAMC-G}
& \multicolumn{2}{c}{NALR-LEBM}
& \multicolumn{2}{c}{DAMC-LEBM} \\
\cmidrule(r){2-3} \cmidrule(r){4-5} 
\cmidrule(r){6-7} \cmidrule(r){8-9} \cmidrule(r){10-11}
{} & 
V. & D. & V. & D. & V. & D. & V. & D. & V. & D.\\
\midrule
MSE & 
0.054 & - & 0.020 & - & 0.018 & 0.015 & 
0.028 & 0.016 & 0.021 & \textbf{0.015} \\
FID & 
78.06 & - & 70.15 & - & 90.30 & 66.93 & 
68.52 & 64.38 & 60.89 & \textbf{57.72} \\
\bottomrule
\end{tabular}
\vspace{-10pt}
\end{table}

\section{Related Work}
\label{sec:relatd_work}
\paragraph{Energy-based prior model}
\acp{ebm} \citep{nijkamp2020anatomy,nijkamp2019learning,xie2016theory,du2019implicit,du2020improved} play an important role in generative modeling. \citet{pang2020learning} propose to learn an \ac{ebm} as a prior model in the latent space of \acp{dlvm}; it greatly improves the model expressivity over those with non-informative priors and brings strong performance on downstream tasks, \eg, image segmentation, text modeling, molecule generation, and trajectory prediction \citep{yu2021unsupervised,yu2022latent,pang2020molecule,pang2021trajectory}. However, learning \acp{ebm} or latent space \acp{ebm} requires \acs{mcmc} sampling to estimate the learning gradients, which needs numerous iterations to converge when the target distributions are high-dimensional or highly multi-modal. Typical choices of sampling with non-convergent short-run MCMC \cite{nijkamp2019learning} in practice can lead to poor generation quality, malformed energy landscapes \cite{yu2022latent,nijkamp2019learning,gao2020learning}, biased estimation of the model parameter and instability in training \citep{nijkamp2020anatomy,gao2020learning,du2019implicit,du2020improved}. In this work, we consider learning valid amortization of the long-run MCMC for energy-based priors; the proposed model shows reliable sampling quality in practice.

\paragraph{Denoising diffusion probabilistic model}
\acp{ddpm} \citep{song2019generative,ho2020denoising,song2020improved}, originating from \cite{sohl2015deep}, learn the generative process by recovering the observed data from a sequence of noise-perturbed versions of the data. The learning objective can be viewed as a variant of the denoising score matching objective \citep{vincent2011connection}. As pointed out in \cite{sohl2015deep,ho2020denoising}, the sampling procedure of \ac{ddpm} with $\be$-prediction parametrization resembles \ac{ld} of an \ac{ebm}; $\be$ (predicted noise) plays a similar role to the gradient of the log density \cite{ho2020denoising}. To be specific, learning a \ac{ddpm} with ${\be}$-prediction parameterization is equivalent to fitting the finite-time marginal of a sampling chain resembling annealed Langevin dynamics \cite{song2019generative,ho2020denoising,song2020improved}. Inspired by this connection, we propose to amortize the long-run MCMC in learning energy-based prior by iteratively distilling the short-run chain segments with a \ac{ddpm}-based sampler. We show empirically and theoretically that the learned sampler is valid for long-run chain sampling.

\paragraph{Amortized MCMC}
The amortized MCMC technique is formally brought up by \citet{li2017approximate}, which incorporates feedback from MCMC back to the parameters of the
amortizer distribution $q_{\bP}$. It is concurrently and independently proposed by \citet{xie2018cooperative} as the MCMC teaching framework. Methods under this umbrella term \cite{li2017approximate,xie2018cooperative,ruiz2019contrastive,xie2020generative,xie2021learning,xie2022tale,zhu2023learning} generally learns the amortizer by minimizing the divergence (typically the \ac{kld}) between the improved distribution and its initialization, i.e., ${\mD}[{\mK}_T q_{{\bP}_{k-1}}||q_{\bP}]$, where ${\mK}_T$ represents $T$-step MCMC transition kernel and $q_{{\bP}_{k-1}}$ represents the current amortizer. The diffusion-based amortization proposed in this work can be viewed as an instantiation of this framework, while our focus is on learning the energy-based prior. Compared with previous methods, our method i) specifically exploits the connection between \acp{ebm} and \acp{ddpm} and is suitable for amortizing the prior and posterior sampling MCMC of energy-based prior, and ii) resides in the lower-dimensional latent space and enables faster sampling and better convergence.

\paragraph{More methods for learning EBM} Several techniques other than short-run MCMC have been proposed to learn the \ac{ebm}. In the seminal work, \citet{hinton2002training} proposes to initialize Markov chains using real data and run several steps of \acs{mcmc} to obtain samples from the model distribution. \citet{tieleman2008training} proposes to start Markov chains from past samples in the previous sampling iteration, known as Persistent Contrastive Divergence (PCD) or persistent chain sampling, to mimic the long-run sampling chain. \citet{nijkamp2020anatomy} provide comprehensive discussions about tuning choices for \ac{ld} such as the step size $s$ and sampling steps $T$ to obtain stable long-run samples for persistent training. \cite{du2019implicit,du2020improved} employ a hybrid of persistent chain sampling and short-run sampling by maintaining a buffer of previous samples. The methods draw from the buffer or initialize the short-run chain with noise distribution with some pre-specified probability. Another branch of work, stemmed from \cite{gutmann2010noise}, considers discriminative contrastive estimation to avoid MCMC sampling. \citet{gao2020flow} use a normalizing flow \cite{rezende2015variational} as the base distribution for contrastive estimation. \citet{aneja2021contrastive} propose to estimate the energy-based prior model based on the prior of a pre-trained VAE \cite{vahdat2020nvae} by noise contrastive estimation. More recently, \citet{xiao2022adaptive} learn a sequence of \acp{ebm} in the latent space with adaptive multi-stage NCE to further improve the expressive power of the model.

\section{Conclusion}
In this paper, we propose the \ac{damc} sampler and develop a novel learning algorithm for \ac{lebm} based on it. We provide theoretical and empirical evidence for the effectiveness of our method. We notice that our method can be applied to amortizing MCMC sampling of unnormalized continuous densities in general. It can also be applied to sampling posterior distributions of continuous latent variables in general latent variable models. We would like to explore these directions in future work.

\section*{Acknowledgements}
Y. N. Wu was supported by NSF
DMS-2015577. We would like to thank the anonymous
reviewers for their constructive comments.

\bibliographystyle{unsrtnat}
\bibliography{reference}

\clearpage
\renewcommand\thefigure{S\arabic{figure}}
\setcounter{figure}{0}
\renewcommand\thetable{S\arabic{table}}
\setcounter{table}{0}
\pagenumbering{arabic}% resets `page` counter to 1
\renewcommand*{\thepage}{S\arabic{page}}
\appendix

\newpage
\section{Theoretical Discussion}
\subsection{Monotonically Decreasing \texorpdfstring{\acs{kld}}{}}
\label{app:mdkld}
We state in the main text that ${\mD} [q_{{\bP}_k} || \pi ] \leq {\mD} [q_{{\bP}_{k-1}} || \pi ]$, where $\pi$ is the stationary distribution. To show this, we first provide a proof of ${\mD} [\pi_{t + T} || \pi ] \leq {\mD} [\pi_{t} || \pi ]$, where $\pi_t$ and $\pi_{t + T}$ are the distributions of $\z$ at $t$-th and $(t+T)$-th iteration, respectively. This is a known result from \cite{cover1999elements}, and we include it here for completeness.
\begin{equation}
\label{equ:dkld}
\begin{aligned}
{\mD} [\pi_{t} || \pi ] &= 
\E_{\pi_t(\z_t)} \left[
\log \frac{\pi_t(\z_t)}{\pi(\z_t)}
\right] = 
\E_{\pi_t(\z_t), {\mK}(\z_{t+1}|z_t)} \left[
\log \frac{\pi_t(\z_t) {\mK}(\z_{t+1}|z_t)}
          {\pi(\z_t) {\mK}(\z_{t+1}|z_t)}
\right] \\ &\stackrel{(i)}{=}
\E_{\pi_{t+1}(\z_{t+1}), {\mK'}_{\pi_{t+1}}(\z_{t}|z_{t+1})} \left[
\log \frac{{\mK'}_{\pi_{t+1}}(\z_{t}|z_{t+1})   
          \pi_{t+1}(\z_{t+1})}
          {{\mK'}_{\pi}(\z_{t}|z_{t+1}) \pi(\z_{t+1})}
\right] \\ &=
{\mD} [\pi_{t+1} || \pi ] + 
\E_{\pi_{t+1}(\z_{t+1})}
{\mD} [{\mK'}_{\pi_{t+1}}(\z_{t}|z_{t+1})  || 
       {\mK'}_{\pi}(\z_{t}|z_{t+1}) ] 
\\ &\stackrel{(ii)}{\ge}
{\mD} [\pi_{t+1} || \pi ],
\end{aligned}
\end{equation}
where we denote the forward-time transition kernel as $\mK$ and the reverse-time kernel as $\mK'$. (i) holds because we are just re-factorizing the joint density of $[\z_t, \z_{t+1}]$: 
$
\pi_t(\z_t) {\mK}(\z_{t+1}|z_t) = 
{\mK'}_{\pi_{t+1}}(\z_{t}|z_{t+1}) 
   \pi_{t+1}(\z_{t+1})
$ and 
$
\pi(\z_t) {\mK}(\z_{t+1}|z_t) = 
{\mK'}_{\pi}(\z_{t}|z_{t+1}) \pi(\z_{t+1})
$. (ii) holds because the \acs{kld} is non-negative. We can see that ${\mD} [\pi_{t + T} || \pi ] \leq {\mD} [\pi_{t} || \pi ]$ is a direct result from \cref{equ:dkld}, and that $\pi_t \to \pi$ as $t \to \infty$ under proper conditions \cite{welling2011bayesian}.

In the main text, we describe the update rule of the sampler $q_{\bP}$ as follows:
\begin{equation}
q_{{\bP}_{k}} \gets 
 \argmin_{q_{\bP} \in {\mQ}} {\mD}[q_{{\bP}_{k-1}, T}||q_{\bP}],~
q_{{\bP}_{k-1}, T} := {\mK}_T q_{{\bP}_{k-1}},~
q_{\bP_0} \approx \pi_0,~k=0,...,K-1.
\end{equation}
In the ideal case, we can assume that the objective in \cref{equ:qphi} is properly optimized and that $\{q_{\bP}\}$ is expressive enough to parameterize each $q_{{\bP}_{k-1}, T}$. With $q_{\bP_0} \approx \pi_0$ and $q_{{\bP}_{k}} \approx {\mK}_T q_{{\bP}_{k-1}}$, we can conclude that ${\mD} [q_{{\bP}_k} || \pi ] \leq {\mD} [q_{{\bP}_{k-1}} || \pi ]$ for each $k = 1,...,K$ according to \cref{equ:dkld}. We will discuss in the following section the scenario where we apply gradient-based methods to minimize ${\mD}[q_{{\bP}_{k-1}, T}||q_{\bP}]$ and approximate \cref{equ:qphi} for the update from $q_{{\bP}_{k-1}}$ to $q_{{\bP}_k}$.

\subsection{Discussion about Diffusion-Based Amortization}
\label{app:diff_amor}
We can see that the statement in \cref{app:mdkld} holds when $q_{{\bP}_{k}}$ is a close approximation of $q_T := {\mK}_T q_{{\bP}_{k-1}}$. This motivates our choice of employing \acs{ddpm} to amortize the \ac{ld} transition, considering its capability of close approximation to the given distribution $\{q_T\}$. Based on the derivation of \ac{ddpm} learning objective in \cite{kingma2013auto}, we know that 
\begin{equation}
\label{equ:cent_app}
\begin{aligned}
\argmin_{q_{\bP}} {\mD}[q_T||q_{\bP}] &=
\argmin_{q_{\bP}} -{\mH}(q_{T}) + {\mH}(q_{T}, q_{\bP}) = 
\argmin_{q_{\bP}} -\E_{q_{T}}\left[
\log q_{\bP}
\right] \\ &\leq 
\argmin_{q_{\bP}} 
\E_{{\be},\lambda}
\left[\|{\be}_{\bP}(\z_\lambda) - {\be} \|_2^2\right]
\approx
\argmin_{q_{\bP}} 
\frac{1}{N} \sum_{j=1}^N
\left[\|{\be}_{\bP}(\z_{j, \lambda_j}) - {\be}_j \|_2^2\right],
\end{aligned}
\end{equation}
where $\z_\lambda$ is draw from $q(\z_\lambda|\z_0) = \N(\z_\lambda; \beta_\lambda \z_0, \sigma_\lambda^2 \I_d)$. $\z_0 \sim q_T$. ${\be} \sim \N({\bZ}, \I_d)$ and $\lambda$ is drawn from a distribution of log noise-to-signal ratio $p(\lambda)$. In practice, we use Monte-Carlo average to approximate the objective and employ a gradient-based update rule for $q_{\bP}$:
\begin{equation}
\label{equ:q_upd_app}
\begin{aligned}
{\bP}_{k-1}^{(i+1)} \gets {\bP}_{k-1}^{(i)} - \eta \nabla_{\bP} 
\frac{1}{N} \sum_{j=1}^N
\left[\|{\be}_{\bP}(\z_{j, \lambda_j}) {\be}_j\|_2^2\right],~
{\bP}_{k}^{(0)} \gets {\bP}_{k-1}^{(M)},~i=0,1,...,M-1.
\end{aligned}
\end{equation}

This can be viewed as a M-estimation of ${\bP}$. Recall that $q_T = {\mK}_T q_{{\bP}_{k-1}}$, we can construct ${\bP}_{k} = \left({\bP}_{k-1}, \tilde{\bP}\right)$ to minimize the \ac{kld}, where $\tilde{\bP}$ models the transition kernel ${\mK}_T$. Therefore, initializing $q_{\bP}$ to be optimized with $q_{{\bP}_{k-1}}$, we are effectively maximizing ${\mL}(\tilde{\bP}) = \frac{1}{N} \sum_{j=1}^N \log p_{\Tilde{\bP}}(\hat{\z}_j | \z_j)$, where $\{\z_j\}$ are from $q_{{\bP}_{k-1}}$ and $\{\hat{\z}_j\}$ are from ${\mK}_T q_{{\bP}_{k-1}}$. 
Let $\Tilde{\bP}$ be the result of M-estimation and $\Tilde{\bP}^*$ be the true target parameter. Then based on the derivation in \cite{gao2020learning}, asymptotically we have
\begin{equation}
\sqrt{N}\left(
\Tilde{\bP} - \Tilde{\bP}^*
\right) \to 
\N\left(0, {\mI}\left(\Tilde{\bP}^*\right)^{-1}\right),
\end{equation}
where 
$
{\mI}\left(\Tilde{\bP}^*\right) = 
\E_{\hat{\z}, \z}
\left[
-\nabla^2 \log p_{\Tilde{\bP}}(\hat{\z} | \z)
\right]
$ is the Fisher information matrix. This interpretation tells us that i) when the sample size $N$ is large, the estimation $\Tilde{\bP}$ is asymptotically unbiased, and ii) if we want to obtain the estimation $\Tilde{\bP}$ with a few gradient-based updates, then the eigenvalues of the Fisher information matrix would be relatively small but non-zero. ii) suggests that ${\mK}_T q_{{\bP}_{k-1}}$ should be significantly different from $q_{{\bP}_{k-1}}$, which is confirmed by \cite{ruiz2019contrastive} and our preliminary experiments, but it should not be too far away because that would require more gradient-based updates. We find that setting $T=30$ and $M=6$ works well in the experiments.

\subsection{Further Discussion about the Learning Algorithm}
\label{app:disc_lalg}
For completeness, we first derive the learning gradients for updating ${\bT}$.
\begin{equation}
\label{equ:mle_app}
\begin{aligned}
\nabla_{\bT} \log p_{\bT}(\x) &= 
\frac{1}{p_{\bT}(\x)} 
\nabla_{\bT} \int_{\z} p_{\bT}(\x, \z) d \z =  
\int_{\z} 
\frac{p_{\bT}(\x, \z)}{p_{\bT}(\x)}
\nabla_{\bT} \log p_{\bT}(\x, \z) d \z \\&= 
\E_{p_{\bT}(\z|\x)}\left[
\nabla_{\bT} \log p_{\bT}(\z, \x)
\right] \\ &= 
(\E_{p_{\bT}(\z|\x)}\left[
\nabla_{\bA} \log p_{\bA}(\z)
\right],
\E_{p_{\bT}(\z|\x)}\left[
\nabla_{\bB} \log p_{\bB}(\x | \z)
\right])
\\
&=(
\underbrace{
\E_{p_{\bT}(\z|\x)} 
    \left[\nabla_{\bA} f_{\bA}(\z) \right] - 
\E_{p_{\bA}(\z)} 
    \left[\nabla_{\bA} f_{\bA}(\z) \right]
}_{\delta_{\bA} (\x)}, 
\underbrace{
\E_{p_{\bT}(\z|\x)} 
    \left[\nabla_{\bB} \log p_{\bB}(\x | \z) \right] 
}_{\delta_{\bB} (\x)}). 
\end{aligned}
\end{equation}
We can see that ${\bT}$ is estimated in a \acs{mle}-by-EM style. The learning gradient is the same as that of directly maximizing the observed data likelihood, while we need to approximate the expectations in \cref{equ:mle_app}. Estimating the expectations is like the E-step, and update ${\bT}$ with \cref{equ:mle_app} is like the M-step in the EM algorithm. The proposed diffusion-based amortization brings better estimation of the expectations in the E-step, and incorporate the feedback from the M-step by running prior and posterior sampling \ac{ld} as follows 
\begin{equation}
\begin{aligned}
\z_{t + 1} &= 
\z_t + 
\frac{s^2}{2} 
\nabla_{\z_t} 
\underbrace{\left(f_{\bA}(\z_t) - \frac{1}{2}\|\z_t\|_2^2\right)}_{\log p_{\bA}(\z_t)} + s \w_t, \\
\z_{t + 1} &= 
\z_t + 
\frac{s^2}{2} 
\nabla_{\z_t} 
\underbrace{\left(-\frac{\|\x - g_{\bB}(\z_t) \|_2^2}{2\sigma^2} + f_{\bA}(\z_t) - \frac{1}{2}\|\z_t\|_2^2\right)}_{\log p_{\bT}(\z|\x) = \log p_{\bT}(\x, \z) + C} 
+ s \w_t,
\end{aligned}
\end{equation}
to obtain training data. Here $t=0,1,...,T$. $\z_0 \sim q_{{\bP}}(\z | \x)$ for posterior sampling and $\z_0 \sim q_{{\bP}}(\z)$ for prior sampling. $\w_t \sim \N({\bZ}, \I_d)$. Note that we plug-in $p_{\bT}(\x, \z)$ for the target distribution of posterior sampling \ac{ld}. This is because given the observed data $\x$, by Bayes' rule we know that $p_{\bT}(\z | \x) \propto p_{\bT}(\x, \z) = p_{\bB}(\x | \z)p_{\bA}(\z)$. The whole learning iteration can be viewed as a variant of the variational EM algorithm \cite{bernardo2003variational}.

\newpage
\section{Network Architecture and Training Details}
\label{app:model_and_arch}
\paragraph{Architecture}
The energy score network $f_{\bA}$ uses a simple fully-connected structure throughout the  experiments. We describe the architecture in details in \cref{tab:en}.
The generator network has a simple deconvolution structure similar to DCGAN \cite{radford2015unsupervised} shown in \cref{tab:gen}.
The denoising diffusion model is implemented by a light-weight MLP-based U-Net \cite{ronneberger2015u} structure (\cref{tab:diff}).
The encoder network to embed the observed images has a fully convolutional structure \cite{long2015fully}, as shown in \cref{tab:enc}.

\begin{table}[!htb]
  \caption{\textbf{Network structures of the energy score network.} LReLU denotes the Leaky ReLU activation function. The slope in Leaky ReLU is set to 0.2. For SVHN and CelebA datasets, we use \texttt{nz=100}. For the CIFAR-10 and CelebA-HQ datasets, we use \texttt{nz=128}. We use \texttt{nz=8} for anomaly detection on the MNIST dataset, and \texttt{nz=7168} for GAN inversion. We use \texttt{ndf=512} for GAN inversion and \texttt{ndf=200} for the rest experiments.}\label{tab:en}
     \centering
     \begin{tabular}{cc}
     \toprule
     {\bf Layers} & {\bf Out Size} \\ 
     \hline
     Input: $\z$ & nz $\in \{8, 100, 128, 7168\}$\\
	 Linear, LReLU & ndf $\in \{200, 512\}$ \\
	 Linear, LReLU & ndf $\in \{200, 512\}$ \\
	 Linear & 1 \\
     \bottomrule
     \end{tabular}
     \vspace{-10pt}
\end{table}

\begin{table}[!htb]
  \caption{\textbf{Network structures of the generator networks} used for the SVHN, CelebA, CIFAR-10, CelebA-HQ and MNIST (from top to bottom) datasets. For GAN inversion, we use the StyleGAN \cite{karras2019style} structure as our generator network. ConvT($n$) indicates a transposed convolutional operation with $n$ output channels. We use \texttt{ngf=64} for the SVHN dataset and \texttt{ngf=128} for the rest. LReLU indicates the Leaky-ReLU activation function. The slope in Leaky ReLU is set to be 0.2.}
  \label{tab:gen}
  \centering
     \begin{tabular}{ccc}
     \toprule
     {\bf Layers} & {\bf Out Size} & {\bf Stride}\\ \hline
     Input: $\z$ & 1x1x100 &- \\
	 4x4 ConvT(ngf x $8$), LReLU & 4x4x(ngf x 8) & 1 \\
	 4x4 ConvT(ngf x $4$), LReLU & 8x8x(ngf x 4) & 2\\
	 4x4 ConvT(ngf x $2$), LReLU & 16x16x(ngf x 2) & 2\\
	 4x4 ConvT(3), Tanh & 32x32x3 & 2 \\ 
     \hline
     {\bf Layers} & {\bf Out Size} & {\bf Stride}\\ \hline
     Input: $\z$ & 1x1x100 &- \\
		4x4 ConvT(ngf x $8$), LReLU & 4x4x(ngf x 8) & 1 \\
		4x4 ConvT(ngf x $4$), LReLU & 8x8x(ngf x 4) & 2\\
		4x4 ConvT(ngf x $2$), LReLU & 16x16x(ngf x 2) & 2\\
		4x4 ConvT(ngf x $1$), LReLU & 32x32x(ngf x 1) & 2\\
		4x4 ConvT(3), Tanh & 64x64x3 & 2 \\ 
     \hline
     {\bf Layers} & {\bf Out Size} & {\bf Stride}\\ \hline
     Input: $\z$ & 1x1x128 & -\\
		8x8 ConvT(ngf x $8$), LReLU & 8x8x(ngf x 8) & 1 \\
		4x4 ConvT(ngf x $4$), LReLU & 16x16x(ngf x 4) & 2\\
		4x4 ConvT(ngf x $2$), LReLU & 32x32x(ngf x 2) & 2\\
		3x3 ConvT(3), Tanh & 32x32x3 & 1 \\ 
     \hline
     {\bf Layers} & {\bf Out Size} & {\bf Stride}\\ \hline
     Input: $\z$ & 1x1x128 &- \\
        4x4 ConvT(ngf x $16$), LReLU & 4x4x(ngf x 16) & 1 \\
	4x4 ConvT(ngf x $8$), LReLU & 8x8x(ngf x 8) 
        & 2 \\
	4x4 ConvT(ngf x $4$), LReLU & 16x16x(ngf x 4) 
        & 2\\
        4x4 ConvT(ngf x $4$), LReLU & 32x32x(ngf x 4) 
        & 2\\
	4x4 ConvT(ngf x $2$), LReLU & 64x64x(ngf x 2)
        & 2\\
	4x4 ConvT(ngf x $1$), LReLU & 128x128x(ngfx1)     & 2\\
	4x4 ConvT(3), Tanh & 256x256x3 & 2 \\ 
     \hline
     {\bf Layers} & {\bf Out Size} & {\bf Stride}\\ \hline
     Input: $\z$ & 1x1x8 & -\\
		7x7 ConvT(ngf x $8$), LReLU & 7x7x(ngf x 8) & 1 \\
		4x4 ConvT(ngf x $4$), LReLU & 14x14x(ngf x 4) & 2\\
		4x4 ConvT(ngf x $2$), LReLU & 28x28x(ngf x 2) & 2\\
		3x3 ConvT(1), Tanh & 28x28x1 & 1 \\ 
     \bottomrule
     \end{tabular}
 \end{table}

\begin{table}[!htb]
    \caption{\textbf{Network structure of the denoising diffusion network.} a) We use the sinusoidal embedding to embed the time index as in \cite{gao2020learning,ho2020denoising}. b) We use the learned fourier feature module \cite{li2021functional} to embed the input $\z$. c) The merged time embedding and context embedding is used to produce a pair of bias and scale terms to shift and scale the embedding of input $\z$. \texttt{nz} is the input dimension, as in \cref{tab:en}. \texttt{nemb} is the dimension of image embedding as in \cref{tab:enc}.}
    \label{tab:diff} 
    \vskip 0.1in
    \centering
    \small
    \begin{tabular}{ccc}
        \toprule
        {\bf Layers} & {\bf Out size} & {\bf Note} \\
        \cmidrule(r){1-1} \cmidrule(r){2-2} \cmidrule(r){3-3}
        
        \multicolumn{3}{c}{\bf Time Embedding} \\
        \hline
        Input: $t$ & $1$ & \tabincell{c}{time index} \\
        Sin. emb.$^{a}$ & $128$ & {} \\
        Linear, SiLU & $128$ & {} \\
        Linear & $128$ & {} \\
        
        \hline
        \multicolumn{3}{c}{\bf Input Embedding} \\
        \hline
        Input: $\z$ & nz & {} \\
        Fr. emb.$^{b}$ & 2x(nz) & {Fourier feature} \\
        
        \hline
        \multicolumn{3}{c}{\bf Basic Block} \\
        \hline
        Input: $\z, \z_{\text{ctx}}, \z_t$ & 
                nzf, nemb, 128 & {$\z$, ctx. and t emb.} \\
        \hline
        Cat, SiLU & nemb + 128 & merge ctx. \& t emb. \\
        Linear, SiLU & nout & {} \\
        \hline
        Linear & nout & Input emb. \\
        Scale-shift$^{c}$ & nout & \tabincell{c}{scale-shift $\z$ emb. \\ w/ merged emb.} \\
        Add $\z$ & nout & \tabincell{c}{
        skip connection \\ from $\z$} \\
        
        \hline
        \multicolumn{3}{c}{\bf Denoising Diffusion Network} \\
        \hline
        Input: $\z, \z_{\text{ctx}}, t$ & nz, nemb, 128 & Input \\
        Embedding & 2x(nz), 128 & \tabincell{c}{Input \& t emb.} \\
        \hline
        Basic Block & 128 & {Encoding} \\
        Basic Block & 256 & {} \\
        Basic Block & 256 & {} \\
        \hline
        Basic Block & 256 & {Intermediate} \\
        \hline
        Basic Block & 256 & {Cat \& Decoding} \\
        Basic Block & 128 & {} \\
        Basic Block & nz & {Output} \\
        \bottomrule
    \end{tabular}
\end{table}

\begin{table}[!htb]
  \caption{\textbf{Network structures of the encoder networks} used for the SVHN, CelebA, CIFAR-10, CelebA-HQ and MNIST (from top to bottom) datasets. For GAN inversion, the encoder network structure is the same as in \cite{zhu2020domain}. Conv($n$)Norm indicates a convolutional operation with $n$ output channels followed by the Instance Normalization \cite{ulyanov2016instance}. We use \texttt{nif=64} and \texttt{nemb=1024} for all the datasets. LReLU indicates the Leaky-ReLU activation function. The slope in Leaky ReLU is set to be 0.2.}
  \label{tab:enc}
  \centering
     \begin{tabular}{ccc}
     \hline
     {\bf Layers} & {\bf Out Size} & {\bf Stride}\\ \hline
     Input: $\x$ & 32x32x3 &- \\
	 3x3 Conv(nif x $1$)Norm, LReLU & 32x32x(nif x 1) & 1 \\
	 4x4 Conv(nif x $2$)Norm, LReLU & 16x16x(nif x 2) & 2\\
	 4x4 Conv(nif x $4$)Norm, LReLU &  8x8x(nif x 4) & 2\\
      4x4 Conv(nif x $8$)Norm, LReLU &  4x4x(nif x
     8) & 2\\
	 4x4 Conv(nemb)Norm, LReLU & 1x1x(nemb) & 1 \\ 
     \hline
      {\bf Layers} & {\bf Out Size} & {\bf Stride}\\ \hline
     Input: $\x$ & 64x64x3 &- \\
	 3x3 Conv(nif x $1$)Norm, LReLU & 64x64x(nif x 1) & 1 \\
	 4x4 Conv(nif x $2$)Norm, LReLU & 32x32x(nif x 2) & 2\\
	 4x4 Conv(nif x $4$)Norm, LReLU & 16x16x(nif x 4) & 2\\
      4x4 Conv(nif x $8$)Norm, LReLU &  8x8x(nif x
     8) & 2\\
      4x4 Conv(nif x $8$)Norm, LReLU &  4x4x(nif x
     8) & 2\\
	 4x4 Conv(nemb)Norm, LReLU & 1x1x(nemb) & 1 \\ 
     \hline
      {\bf Layers} & {\bf Out Size} & {\bf Stride}\\ \hline
     Input: $\x$ & 32x32x3 &- \\
	 3x3 Conv(nif x $1$)Norm, LReLU & 32x32x(nif x 1) & 1 \\
	 4x4 Conv(nif x $2$)Norm, LReLU & 16x16x(nif x 2) & 2\\
	 4x4 Conv(nif x $4$)Norm, LReLU &  8x8x(nif x 4) & 2\\
      4x4 Conv(nif x $8$)Norm, LReLU &  4x4x(nif x
     8) & 2\\
	 4x4 Conv(nemb)Norm, LReLU & 1x1x(nemb) & 1 \\
     \hline
     {\bf Layers} & {\bf Out Size} & {\bf Stride}\\ \hline
     Input: $\x$ & 256x256x3 &- \\
	 3x3 Conv(nif x $1$)Norm, LReLU & 256x256x(nif x 1) & 1 \\
	 4x4 Conv(nif x $2$)Norm, LReLU & 128x128x(nif x 2) & 2\\
	 4x4 Conv(nif x $4$)Norm, LReLU &  64x64x(nif x 4) & 2\\
      4x4 Conv(nif x $4$)Norm, LReLU &  32x32x(nif x 4) & 2\\
      4x4 Conv(nif x $8$)Norm, LReLU &  16x16x(nif x 8) & 2\\
      4x4 Conv(nif x $8$)Norm, LReLU &  8x8x(nif x
      8) & 2\\
      4x4 Conv(nif x $8$)Norm, LReLU &  4x4x(nif x
      8) & 2\\
	 4x4 Conv(nemb)Norm, LReLU & 1x1x(nemb) & 1 \\
     \hline
     {\bf Layers} & {\bf Out Size} & {\bf Stride}\\ \hline
     Input: $\x$ & 28x28x3 &- \\
	 3x3 Conv(nif x $1$)Norm, LReLU & 28x28x(nif x 1) & 1 \\
	 4x4 Conv(nif x $2$)Norm, LReLU & 14x14x(nif x 2) & 2\\
	 4x4 Conv(nif x $4$)Norm, LReLU &  7x7x(nif x 4) & 2\\
      4x4 Conv(nif x $8$)Norm, LReLU &  3x3x(nif x
     8) & 2\\
	 3x3 Conv(nemb)Norm, LReLU & 1x1x(nemb) & 1 \\
     \bottomrule
     \end{tabular}
 \end{table}

\paragraph{Hyperparameters and training details}
As mentioned in the main text, for the posterior and prior \acs{damc} samplers, we set the number of diffusion steps to $100$. The number of iterations in \cref{equ:q_upd} is set to $M=6$ for the experiments. The \ac{ld} runs $T=30$ and $T=60$ iterations for posterior and prior updates during training with a step size of $s=0.1$. For test time sampling from ${\mK}_{T, \z_i | \x_i} q_{{\bP}_{k}}(\z_i | \x_i)$, we set $T=10$ for the additional \ac{ld}. For test time prior sampling of \acs{lebm} with \ac{ld}, we follow \cite{pang2020learning,xiao2022adaptive} and set $T=100$. To further stabilize the training procedure, we i) perform gradient clipping by setting the maximal gradient norm as $100$, ii) use a separate target diffusion network which is the EMA of the current diffusion network to initialize the prior and posterior updates and iii) add noise-initialized prior samples for the prior updates. These set-ups are identical across different datasets. 

The parameters of all the networks are initialized with the default pytorch methods \cite{paszke2019pytorch}. We use the Adam optimizer \cite{kingma2014adam} with $\beta_1=0.5$ and $\beta_2=0.999$ to train the generator network and the energy score network. We use the AdamW optimizer \cite{loshchilov2017decoupled} with $\beta_1=0.5$, $\beta_2=0.999$ and \texttt{weight\_decay=1e-4} to train the diffusion network. The initial learning rates of the generator and diffusion networks are \texttt{2e-4}, and \texttt{1e-4} for the energy score network. The learning rates are decayed with a factor of $0.99$ every 1K training iterations, with a minimum learning rate of \texttt{1e-5}. We run the experiments on a A6000 GPU with the batch size
of $128$. For GAN inversion, we reduce the batch size to 64. Training typically converges within 200K iterations on all the datasets.

\clearpage
\newpage
\section{Pytorch-style Pseudocode}
\label{app:pcode}
We provide pytorch-style pseudocode to help understand the proposed method. We denote the generator network as \texttt{G}, the energy score network as \texttt{E} and the diffusion network as \texttt{Q}. The first page sketches the prior and posterior sampling process. The second page outlines the learning procedure.

\begin{lstlisting}[language=Python, caption={Prior and posterior LD sampling.}]
def sample_langevin_prior_z(z, netE):
    s = step_size

    for i in range(n_steps):
        en = netE(z).sum()
        z_norm = 1.0 / 2.0 * torch.sum(z**2)
        z_grad = torch.autograd.grad(en + z_norm, z)[0]
        w = torch.randn_like(z)

        # Prior LD Update
        z.data = z.data - 0.5 * (s ** 2) * z_grad + s * w
        
    return z.detach()

def sample_langevin_posterior_z(z, x, netG, netE):
    s = step_size
    sigma_inv = 1.0 / (2.0 * sigma ** 2)

    for i in range(n_steps):
        x_hat = netG(z)
        g_log_lkhd = sigma_inv * torch.sum((x_hat - x) ** 2)
        
        z_n = 1.0 / 2.0 * torch.sum(z**2) 
        en = netE(z).sum()
        
        total_en = g_log_lkhd + en + z_n
        z_grad = torch.autograd.grad(total_en, z)[0]
        w = torch.randn_like(z)

        # Posterior LD Update
        z.data = z.data - 0.5 * (s ** 2) * z_grad + s * w
    return z.detach()
\end{lstlisting}

\newpage
\begin{lstlisting}[language=Python, caption={Learning LEBM with DAMC.}]
for x in dataset:
    # mask for unconditional learning of DAMC
    z_mask_prob = torch.rand((len(x),), device=x.device)
    z_mask = torch.ones(len(x), device=x.device)
    z_mask[z_mask_prob < 0.2] = 0.0
    z_mask = z_mask.unsqueeze(-1)

    # draw DAMC samples
    z0 = Q(x)
    zk_pos, zk_neg = z0.detach().clone(), z0.detach().clone()

    # prior and posterior updates
    zk_pos = sample_langevin_posterior_z(
            z=zk_pos, x=x, netG=G, netE=E)
    zk_neg = sample_langevin_prior_z(
            z=torch.cat(
            [zk_neg, torch.randn_like(zk_neg)], dim=0), 
            netE=E)

    # update Q 
    for __ in range(6):
        Q_optimizer.zero_grad()
        Q_loss = Q.calculate_loss(
        x=x, z=zk_pos, mask=z_mask).mean()
        Q_loss.backward()
        Q_optimizer.step()
        
    # update G
    G_optimizer.zero_grad()
    x_hat = G(zk_pos)
    g_loss = torch.sum((x_hat - x) ** 2, dim=[1,2,3]).mean()
    g_loss.backward()
    G_optimizer.step()

    # update E
    E_optimizer.zero_grad()
    e_pos, e_neg = E(zk_pos), E(zk_neg)
    E_loss = e_pos.mean() - e_neg.mean()
    E_loss.backward()
    E_optimizer.step()
\end{lstlisting}

\newpage
\section{Dataset and Experiment Settings}
\label{app:data_n_expr}
\paragraph{Datasets}
We include the following datasets to study our method: SVHN (32 × 32 × 3), CIFAR-10 (32 × 32 × 3), CelebA (64 × 64 × 3), CeleAMask-HQ (256 x 256 x 3) and MNIST (28 x 28 x 1). Following \citet{pang2020learning}, we use the full training set of SVHN (73,257) and CIFAR-10 (50,000), and take 40,000 samples of CelebA as the training data. We take 29,500 samples from the CelebAMask-HQ dataset as the training data, and test the model on 500 held-out samples. For anomaly detection on MNIST dataset, we follow the experimental settings in \cite{pang2020learning,xiao2022adaptive,kumar2019maximum,zenati2018efficient} and use 80\% of the in-domain data to train the model. The images are scaled to $[-1, 1]$ and randomly horizontally flipped with a prob. of .5 for training. 

\paragraph{GAN inversion settings}
We attempt to use the \acs{damc} sampler for GAN version on the FFHQ (256 x 256 x 3) and LSUN-Tower (256 x 256 x 3) datasets. We take 69,500 samples from the CelebAMask-HQ dataset as the training data, and use the held-out 500 samples for testing. We follow the default data splits of the LSUN dataset. 

For the \acs{lebm}-based inversion method, we train a \acs{lebm} in the 14 x 512 = 7168 dim. latent space. During training, we add $l_2$-regularization on the energy score of \acs{lebm} to stabilize training, as suggested in \cite{du2019implicit}. For the \acs{damc} sampler and the encoder-based inversion method \cite{zhu2016generative}, we generate initial posterior samples of the training data using \cite{zhu2020domain} and train the \acs{damc} sampler and the encoder-based method for 5K iterations using these samples as a warm-up step. The encoder-based method is trained by minimizing the $l_2$ distance between the encoder output and the target samples. After that, these two methods are trained with the default learning algorithms.

\newpage
\section{Additional Qualitative Results}
\label{app:quali_res}
\subsection{Toy Examples}
\label{app:toy}

\begin{figure}[!htbp]
    \centering    \includegraphics[width=0.1325\textwidth]{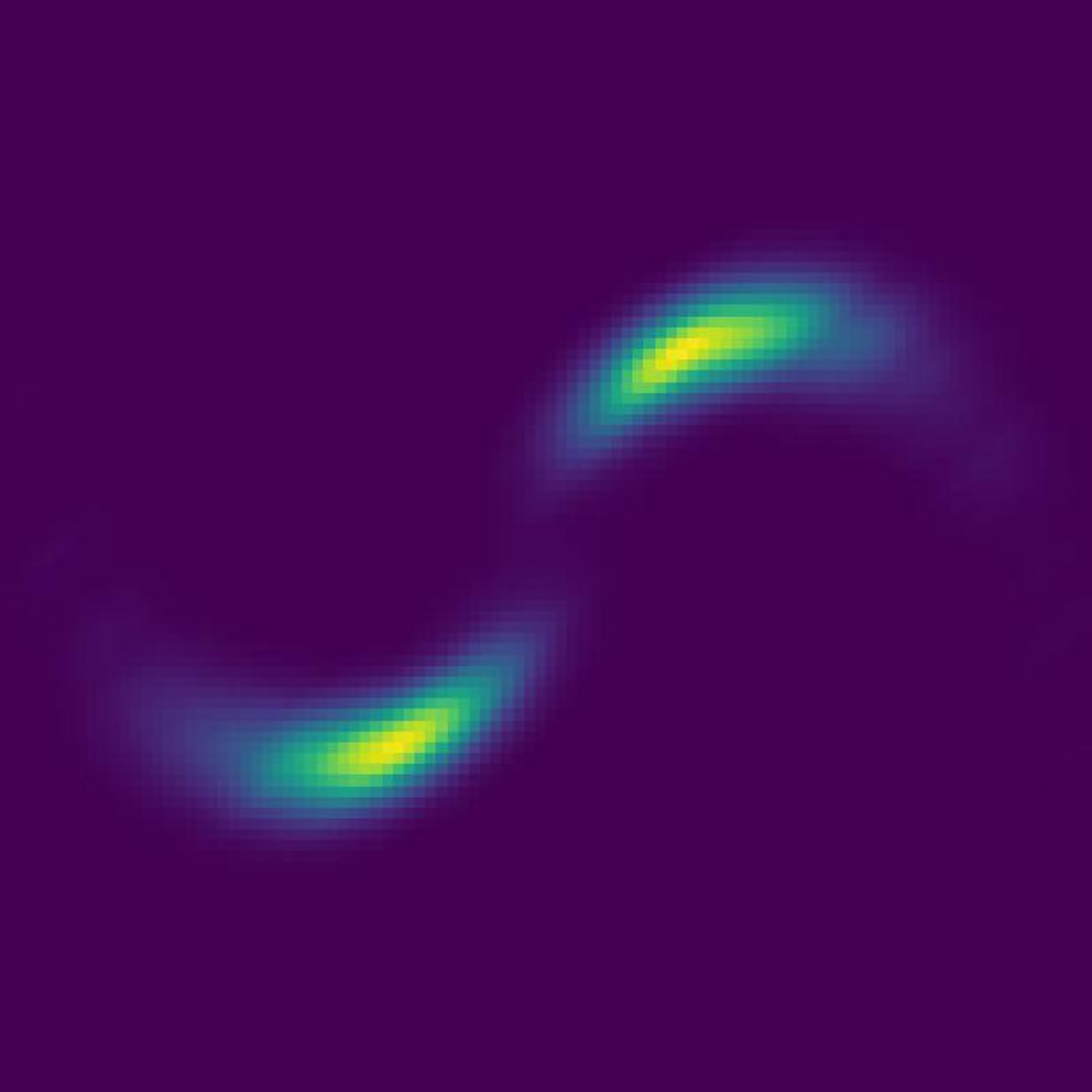}
    \caption{\textbf{The 2-arm pinwheel-shaped prior distribution used in the toy example.}}
    \label{fig:prior_app}
\end{figure}

\begin{figure}[!htbp]
    \centering
    \subfigure[Evolution of the learned posterior distributions]{
        \includegraphics[width=0.80\textwidth]{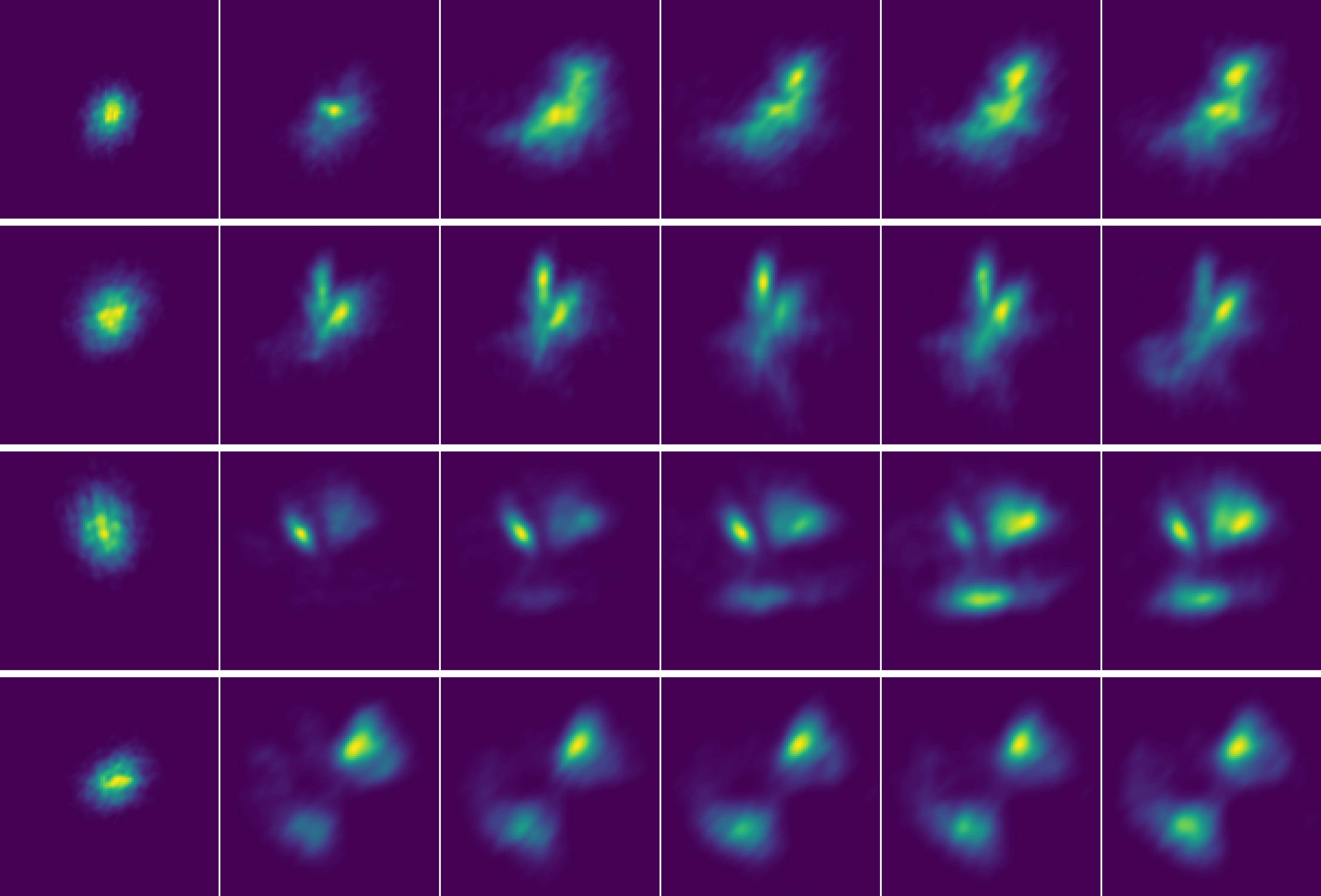}
    }
    \subfigure[GT]{
        \includegraphics[width=0.1325\textwidth]{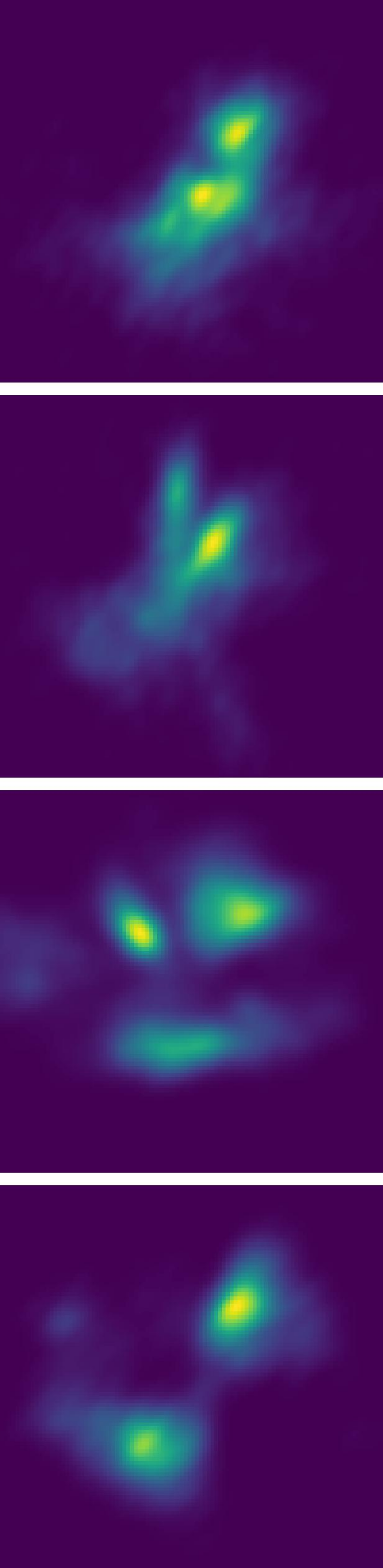}
    }
    \caption{\textbf{Evolution of the posterior distributions learned by the DAMC sampler.} In each row, we display from left to right the evolution of the learned posterior distributions from the DAMC sampler through training iterations. The last column shows the corresponding ground-truth posterior distribution obtained by running 1K-3K steps of Langevin Dynamics until convergence for posterior sampling.}
    \label{fig:post_app}
\end{figure}

As the proof-of-concept toy examples, we implement the neural likelihood experiments following the same set-up mentioned in Section 5.1 and B.1 in \citep{taniguchi2022langevin}. We choose to use a more complex prior distribution, i.e., 2-arm pinwheel-shaped prior distribution (shown in \cref{fig:prior_app}) instead of a standard normal one to make sure that the true posterior distributions are multimodal. The ground-truth posterior distributions are obtained by performing long-run \ac{ld} sampling until convergence. We visualize the convergence trajectory of the posterior distributions learned by our model to the ground-truth ones. In \cref{fig:post_app}, we can see that our model can faithfully reproduce the ground-truth distributions with sufficient training iterations, which indicates that the learned sampler successfully amortizes the long-run sampling chain with the length of $1,000$-$3,000$ iterations.

\newpage
\subsection{Generation} 
\label{app:gen_app}
\begin{figure}[!htbp]
    \centering
    \subfigure[Ours-DAMC]{
        \includegraphics[width=0.47\textwidth]{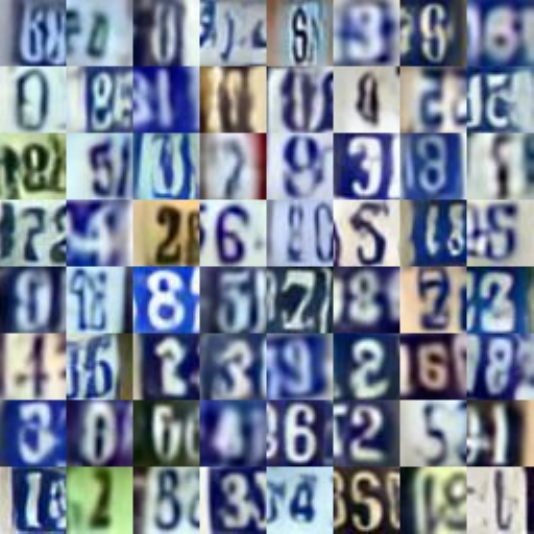}
    }
    \quad
    \subfigure[Ours-LEBM]{
        \includegraphics[width=0.47\textwidth]{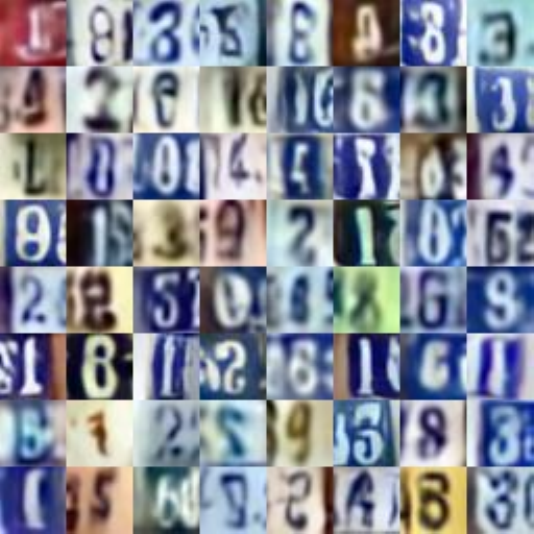}
    }
    \caption{\textbf{Samples generated from the \acs{damc} sampler and \acs{lebm}} trained on the  SVHN dataset.}
    \label{fig:gen_svhn}
    \vspace{-10pt}
\end{figure}

\begin{figure}[!htbp]
    \centering
    \subfigure[Ours-DAMC]{
        \includegraphics[width=0.47\textwidth]{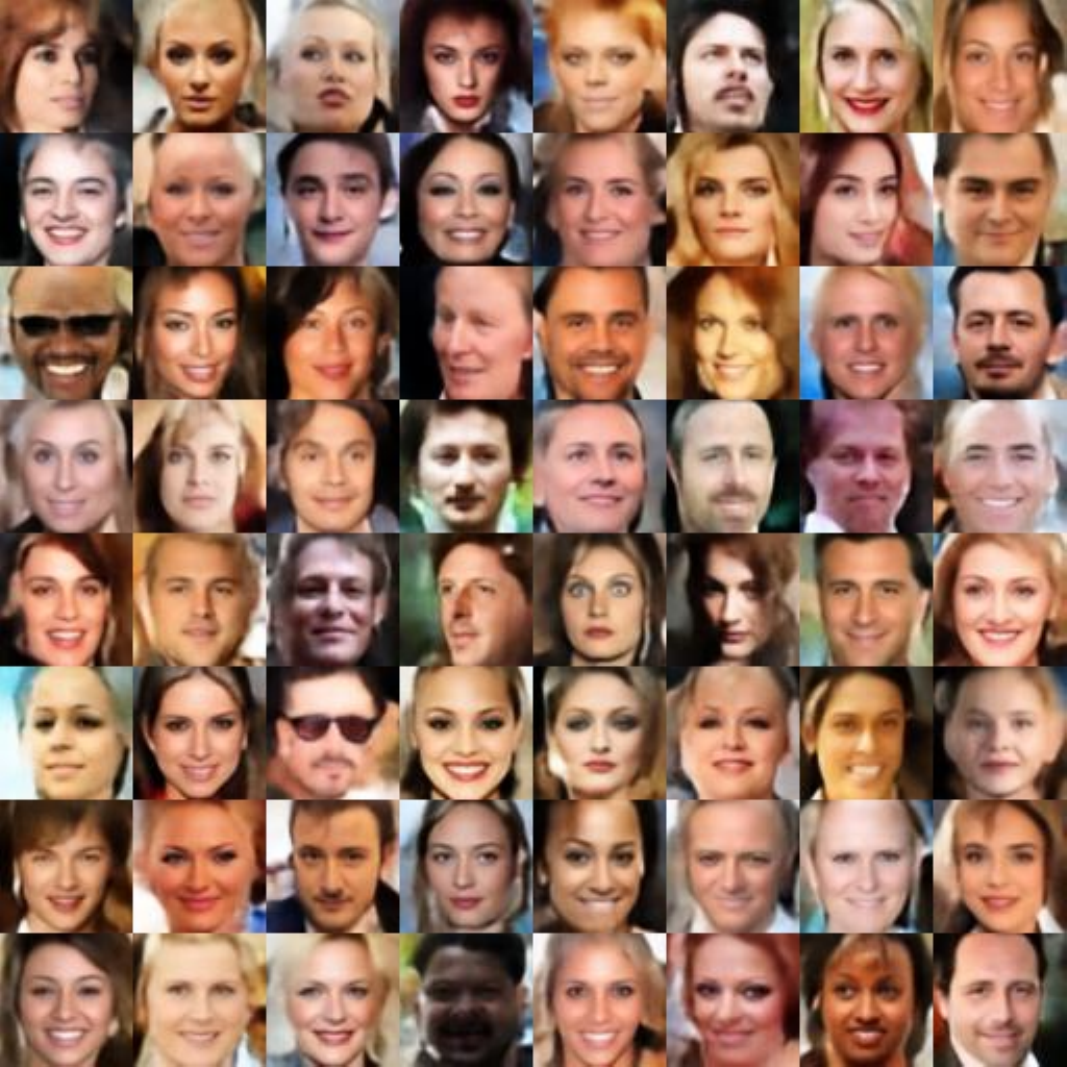}
    }
    \quad
    \subfigure[Ours-LEBM]{
        \includegraphics[width=0.47\textwidth]{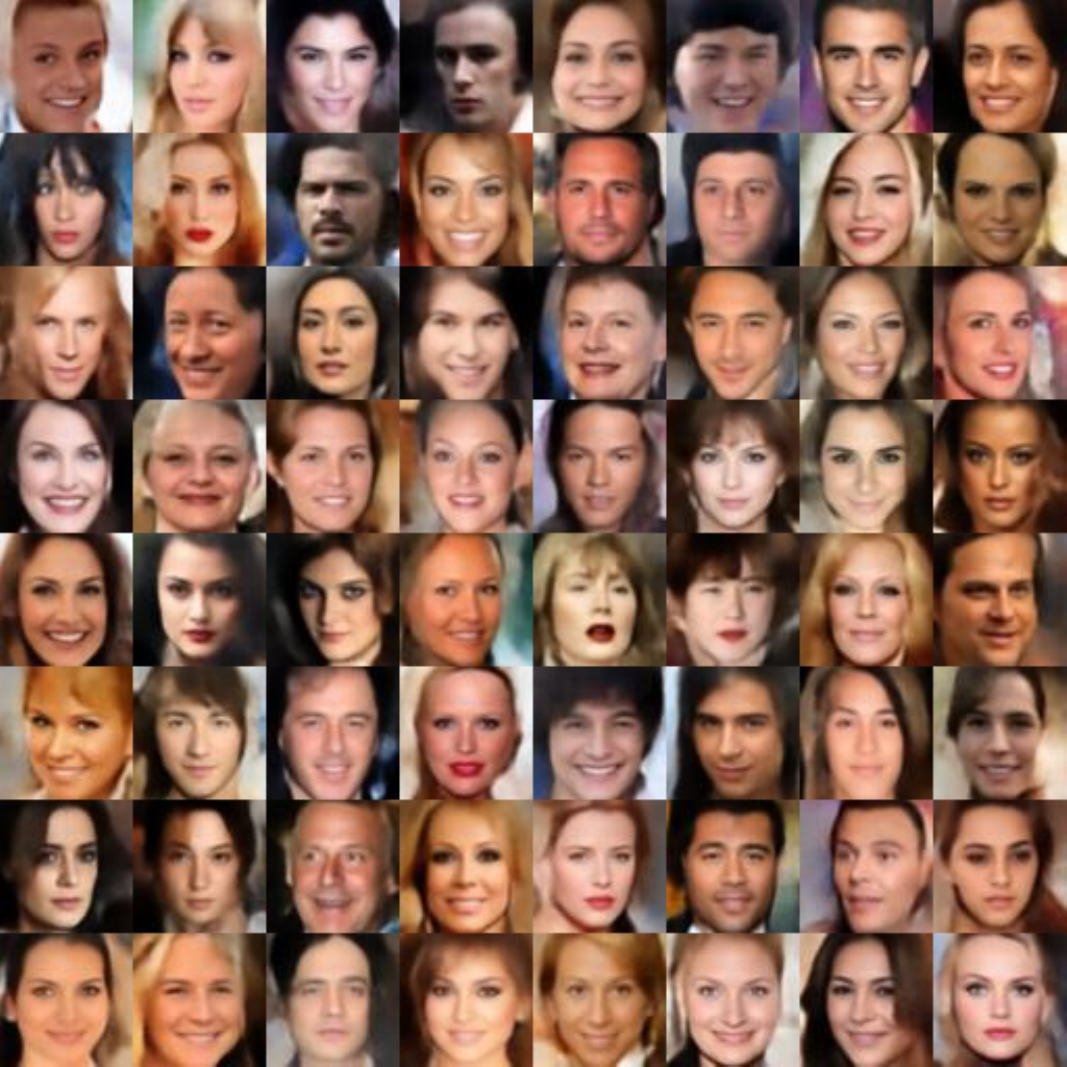}
    }
    \caption{\textbf{Samples generated from the \acs{damc} sampler and \acs{lebm}} trained on the  CelebA dataset.}
    \label{fig:gen_celeba}
    \vspace{-10pt}
\end{figure}

\begin{figure}[!htbp]
    \centering
    \subfigure[Ours-DAMC]{
        \includegraphics[width=0.47\textwidth]{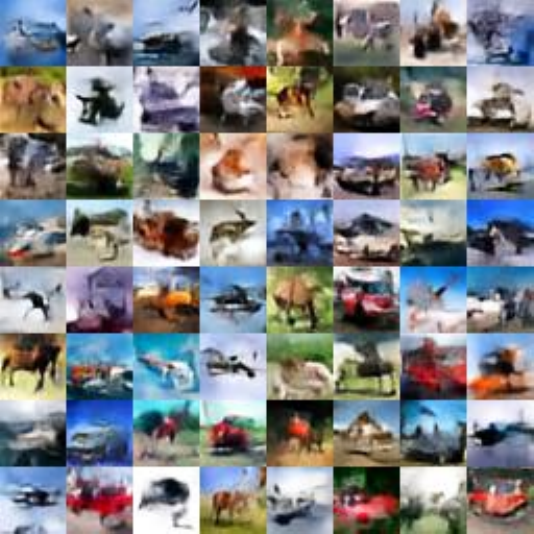}
    }
    \quad
    \subfigure[Ours-LEBM]{
        \includegraphics[width=0.47\textwidth]{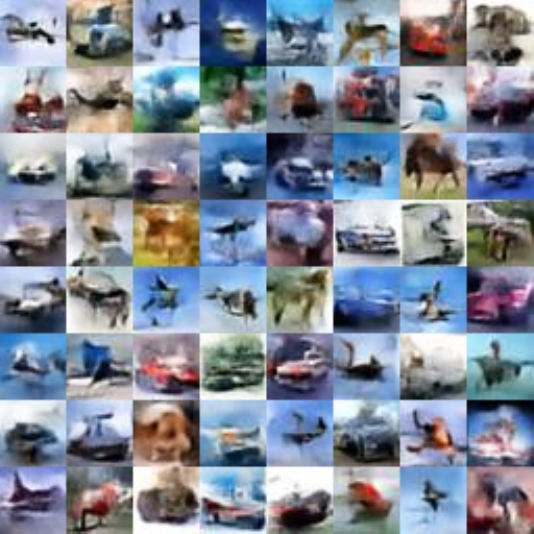}
    }
    \caption{\textbf{Samples generated from the \acs{damc} sampler and \acs{lebm}} trained on the  CIFAR-10 dataset.}
    \label{fig:gen_cifar}
    \vspace{-10pt}
\end{figure}

\begin{figure}[!htbp]
    \centering
    \subfigure[Ours-DAMC]{
        \includegraphics[width=0.47\textwidth]{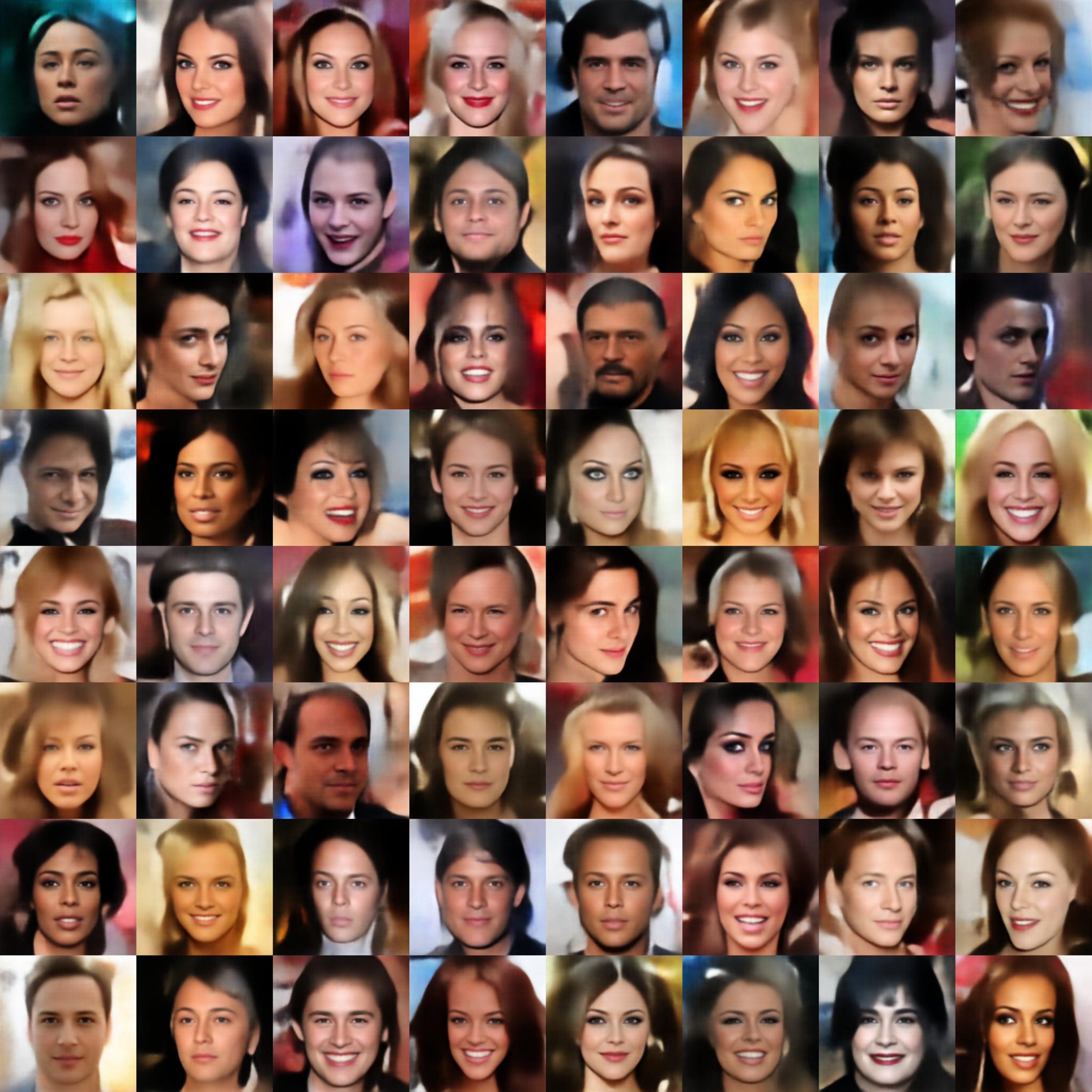}
    }
    \quad
    \subfigure[Ours-LEBM]{
        \includegraphics[width=0.47\textwidth]{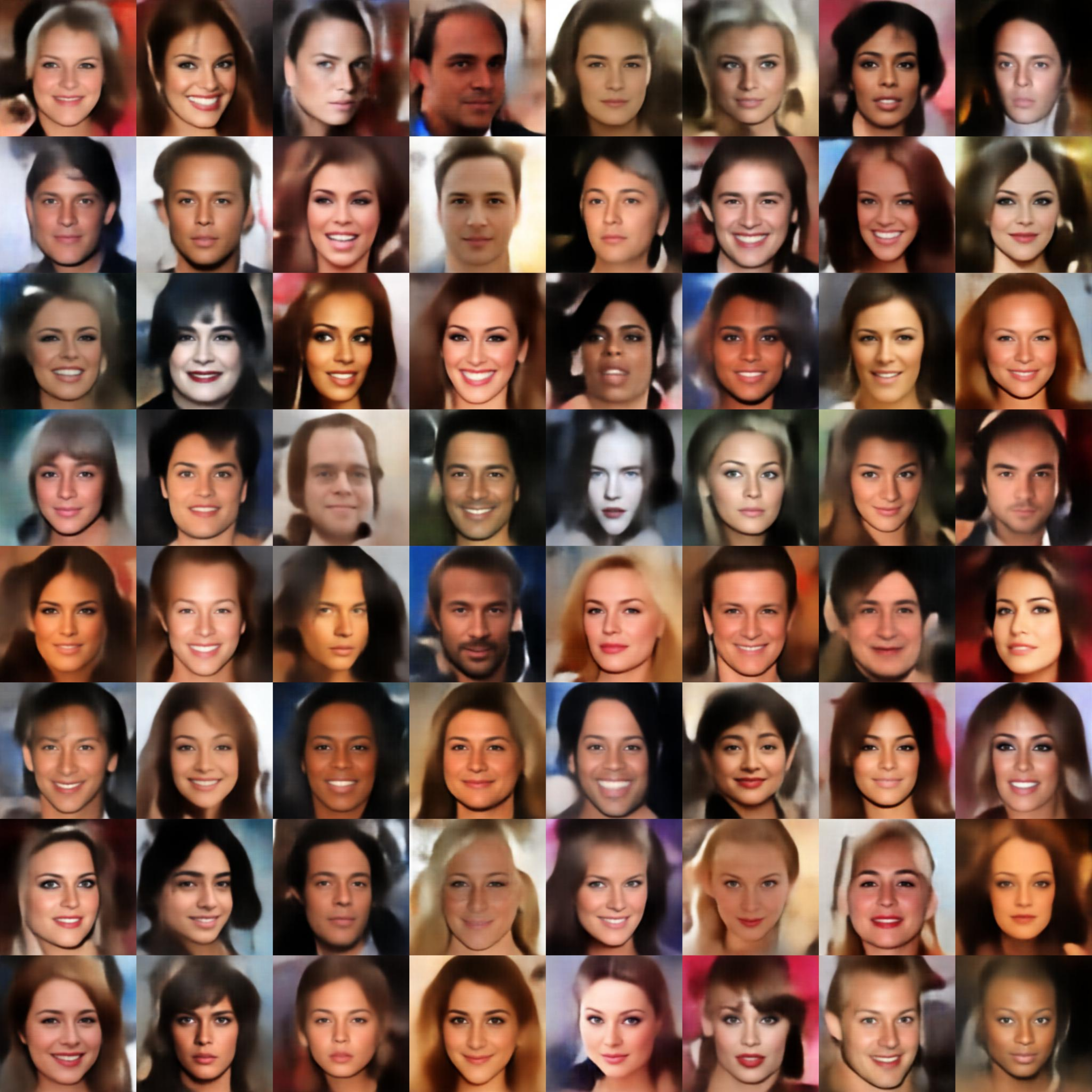}
    }
    \caption{\textbf{Samples generated from the \acs{damc} sampler and \acs{lebm}} trained on the  CelebA-HQ dataset.}
    \label{fig:gen_hq}
    \vspace{-10pt}
\end{figure}

We provide additional generated samples from our models trained on SVHN (\cref{fig:gen_svhn}), CelebA (\cref{fig:gen_celeba}), CIFAR-10 (\cref{fig:gen_cifar}) and CelebA-HQ (\cref{fig:gen_hq}).

\subsection{Reconstruction}
\label{app:recon_app}
\begin{figure}[!htbp]
    \centering
    \subfigure[Observation]{
        \includegraphics[width=0.47\textwidth]{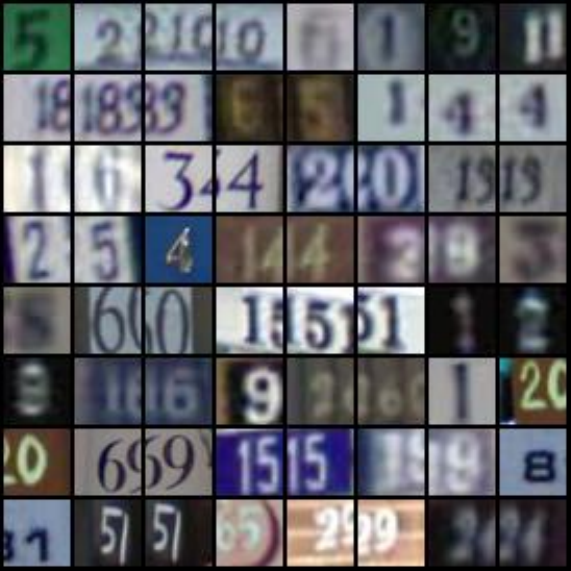}
    }
    \quad
    \subfigure[Reconstruction]{
        \includegraphics[width=0.47\textwidth]{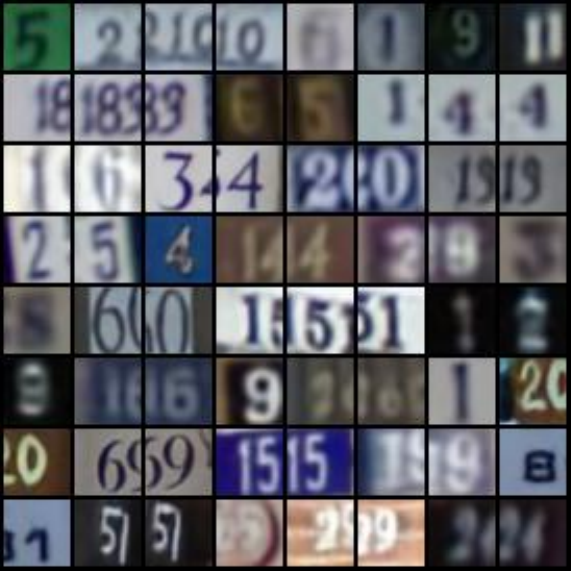}
    }
    \caption{\textbf{Reconstructed samples from the posterior \acs{damc} sampler} trained on the SVHN dataset.}
    \label{fig:rec_svhn}
    \vspace{-10pt}
\end{figure}

\begin{figure}[!htbp]
    \centering
    \subfigure[Observation]{
        \includegraphics[width=0.47\textwidth]{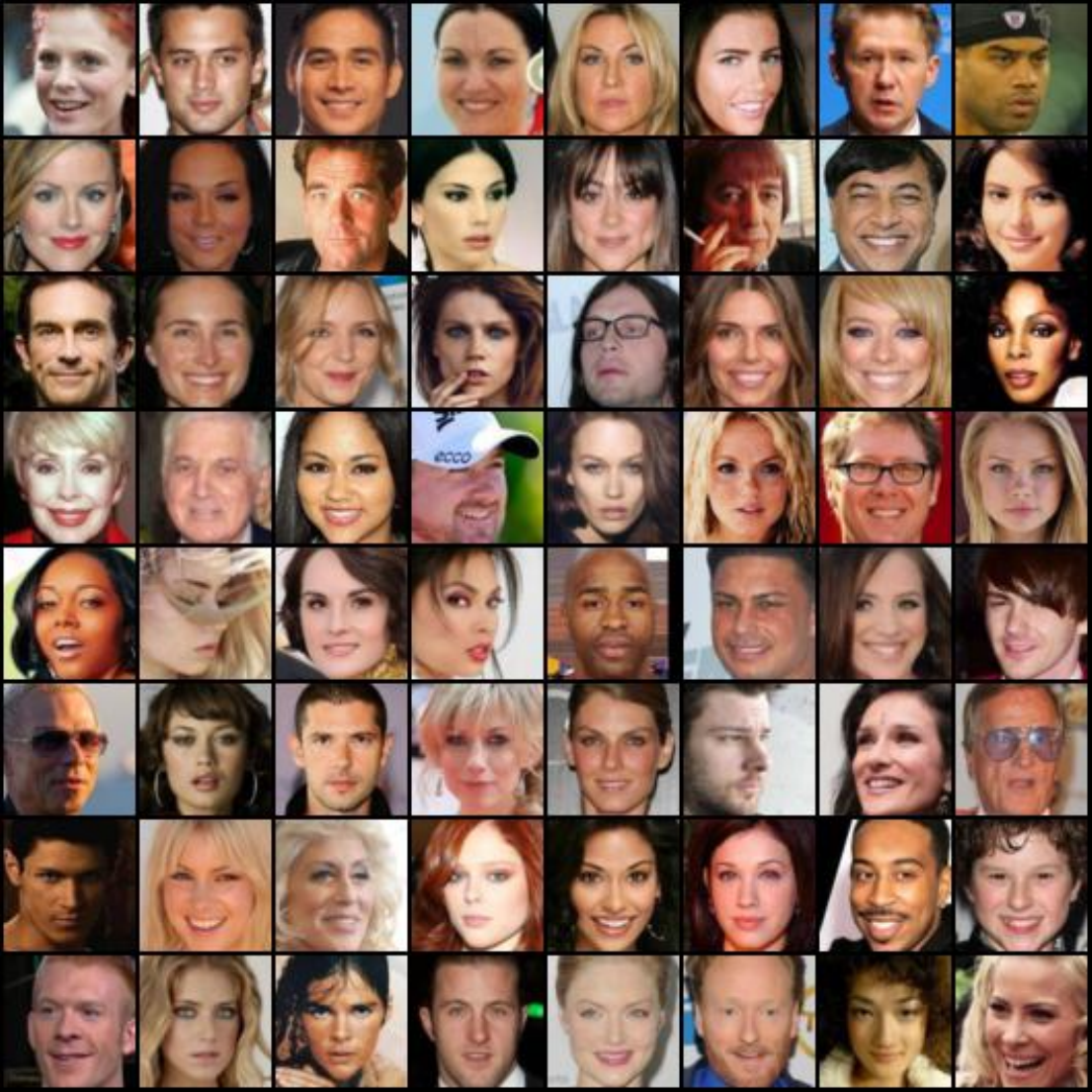}
    }
    \quad
    \subfigure[Reconstruction]{
        \includegraphics[width=0.47\textwidth]{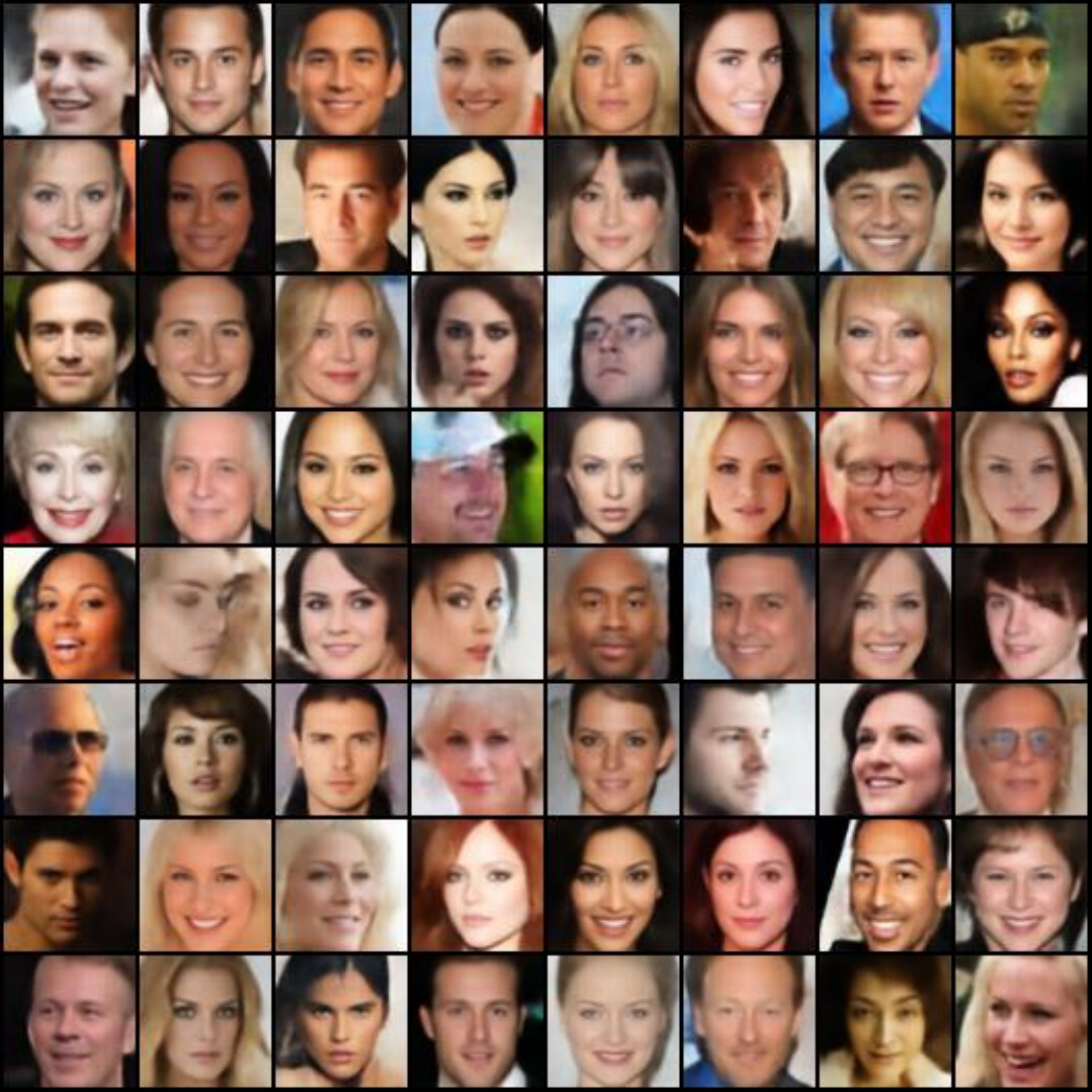}
    }
    \caption{\textbf{Reconstructed samples from the posterior \acs{damc} sampler} trained on the CelebA dataset.}
    \label{fig:rec_celeba}
    \vspace{-10pt}
\end{figure}

\begin{figure}[!htbp]
    \centering
    \subfigure[Observation]{
        \includegraphics[width=0.47\textwidth]{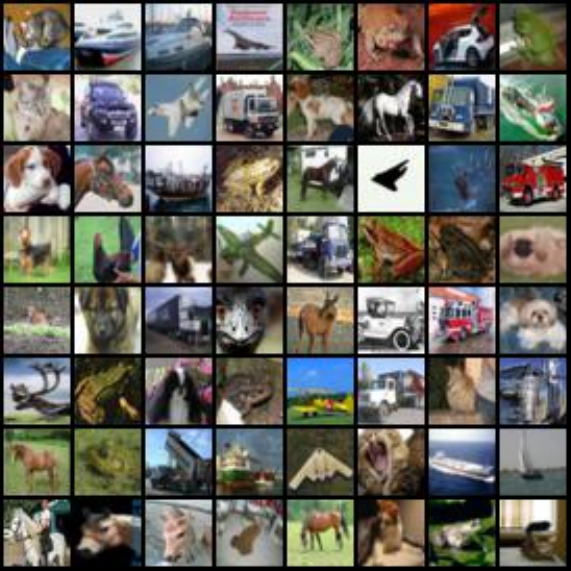}
    }
    \quad
    \subfigure[Reconstruction]{
        \includegraphics[width=0.47\textwidth]{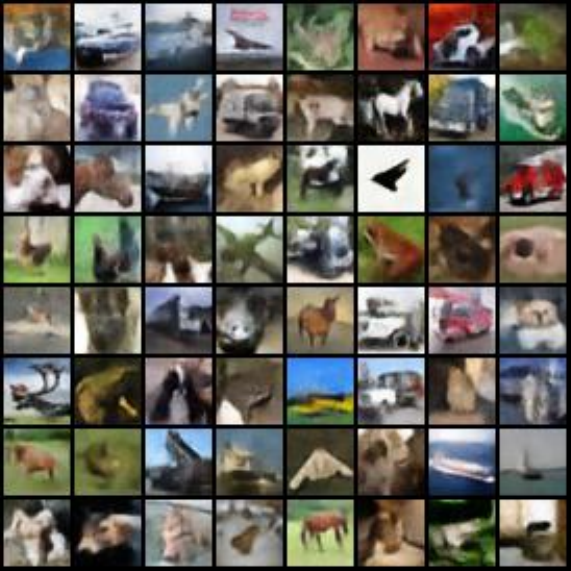}
    }
    \caption{\textbf{Reconstructed samples from the posterior \acs{damc} sampler} trained on the CIFAR-10 dataset.}
    \label{fig:rec_cifar}
    \vspace{-10pt}
\end{figure}

\begin{figure}[!htbp]
    \centering
    \subfigure[Observation]{
        \includegraphics[width=0.47\textwidth]{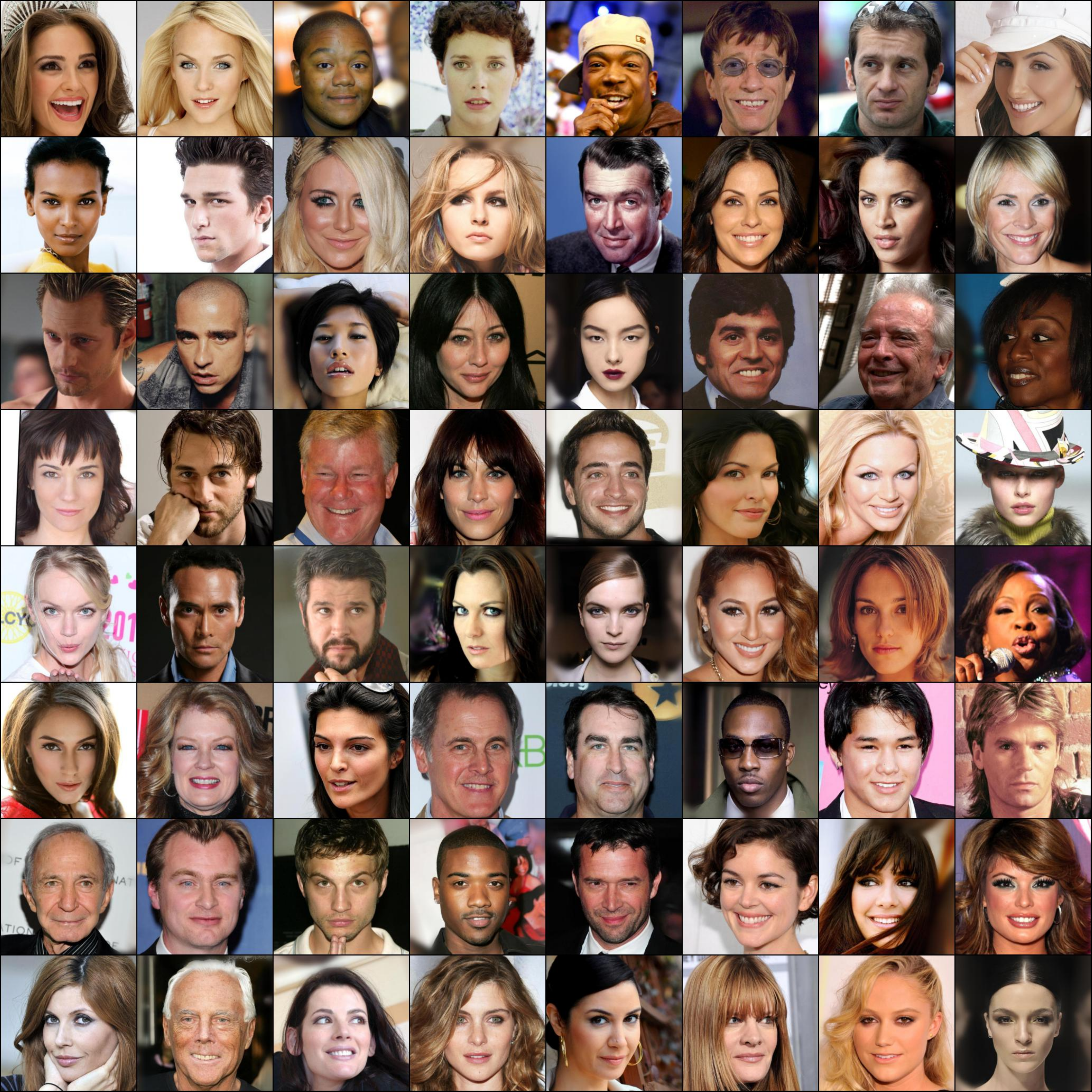}
    }
    \quad
    \subfigure[Reconstruction]{
        \includegraphics[width=0.47\textwidth]{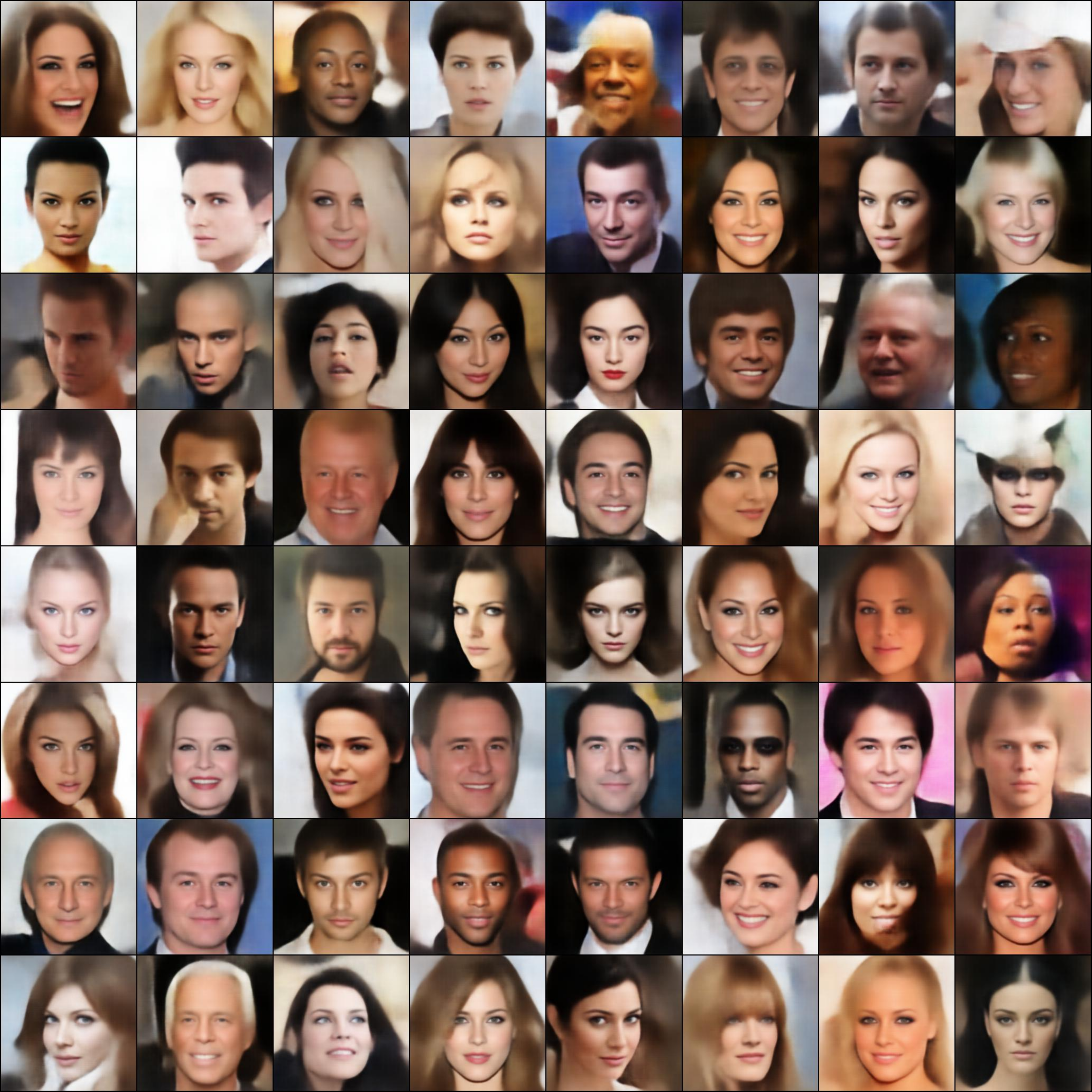}
    }
    \caption{\textbf{Reconstructed samples from the posterior \acs{damc} sampler} trained on the CelebA-HQ dataset.}
    \label{fig:rec_hq}
    \vspace{-10pt}
\end{figure}

We provide qualitative examples about the reconstruction results from our models trained on SVHN (\cref{fig:rec_svhn}), CelebA (\cref{fig:rec_celeba}), CIFAR-10 (\cref{fig:rec_cifar}) and CelebA-HQ (\cref{fig:rec_hq}). Observed images are sampled from the testing set unseen during training.

\newpage
\subsection{Visualization of Transitions}
\label{app:trans}
\begin{figure}[!htbp]
    \centering
    \subfigure[200 steps]{
        \includegraphics[width=0.47\textwidth]{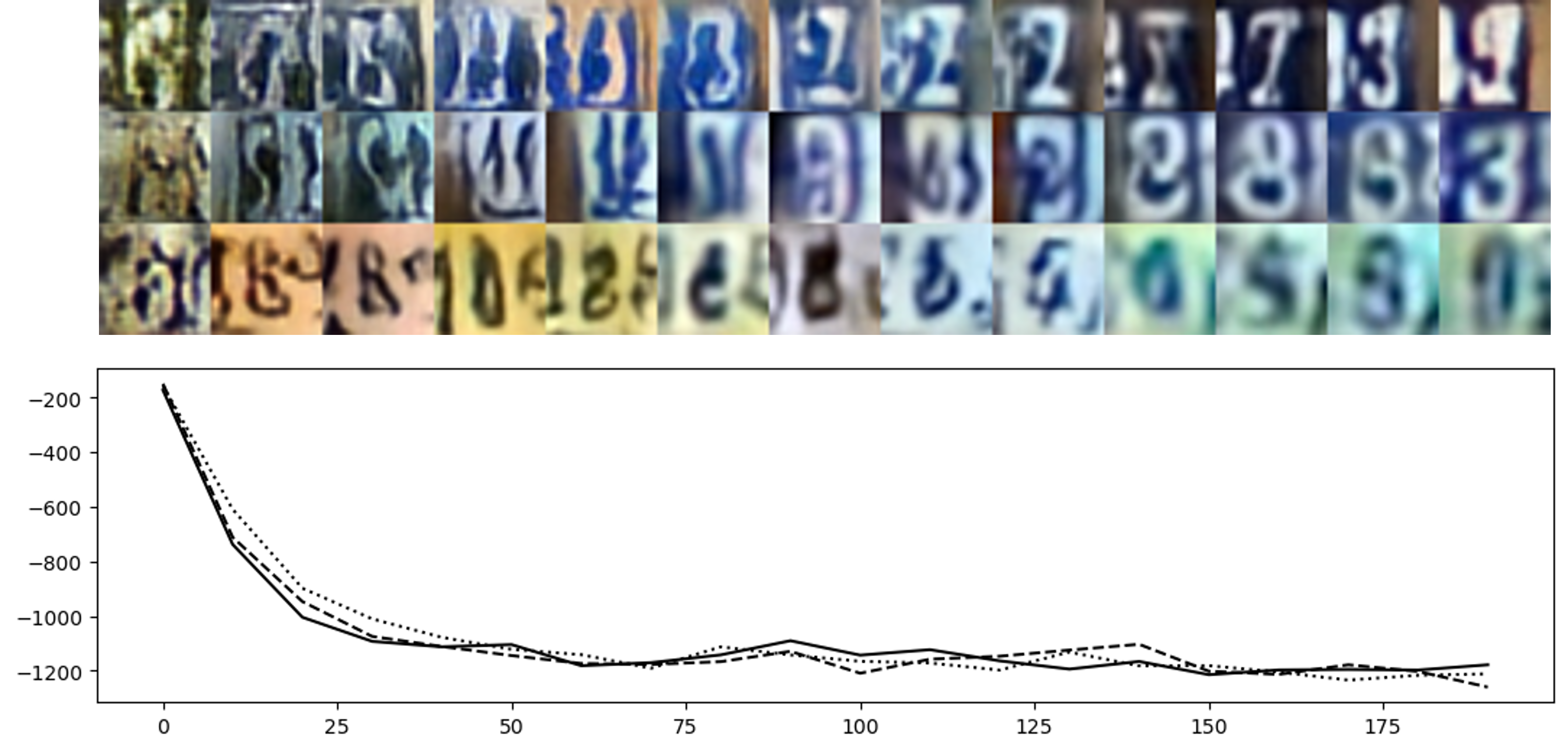}
    }
    \subfigure[2500 steps]{
        \includegraphics[width=0.475\textwidth]{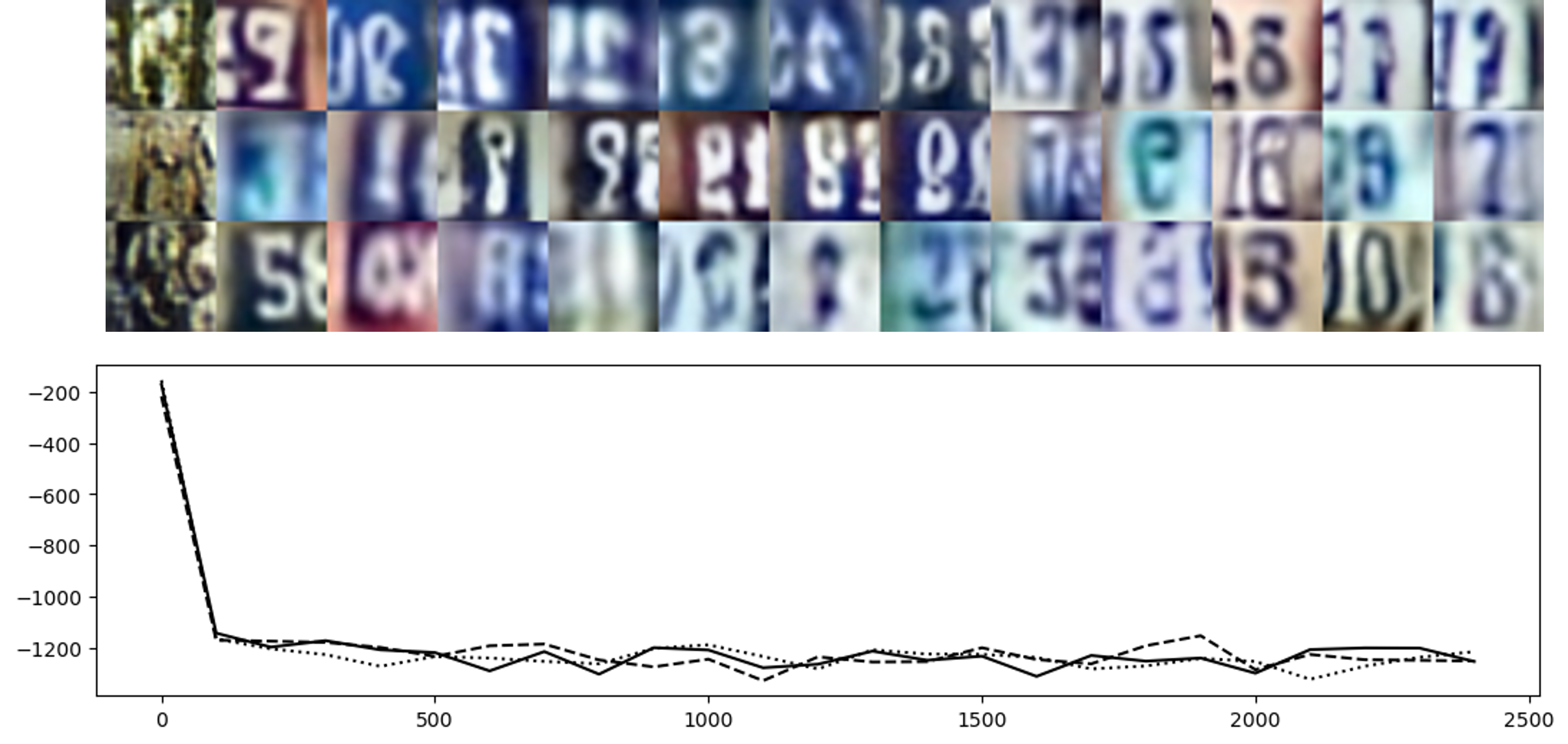}
    }
    \caption{\textbf{Transition of Markov chains initialized from $\N(0, \I_d)$ towards $p_{\bA}(\z)$ on SVHN.} We present results by running \ac{ld} for 200 and 2500 steps. In each sub-figure, the top panel displays the trajectory in the data space uniformly sampled along the chain. The bottom panel shows the energy score $f_{\bA}(\z)$ over the iterations.}
    \label{fig:anl_svhn}
    \vspace{-10pt}
\end{figure}

\begin{figure}[!htbp]
    \centering
    \subfigure[200 steps]{
        \includegraphics[width=0.47\textwidth]{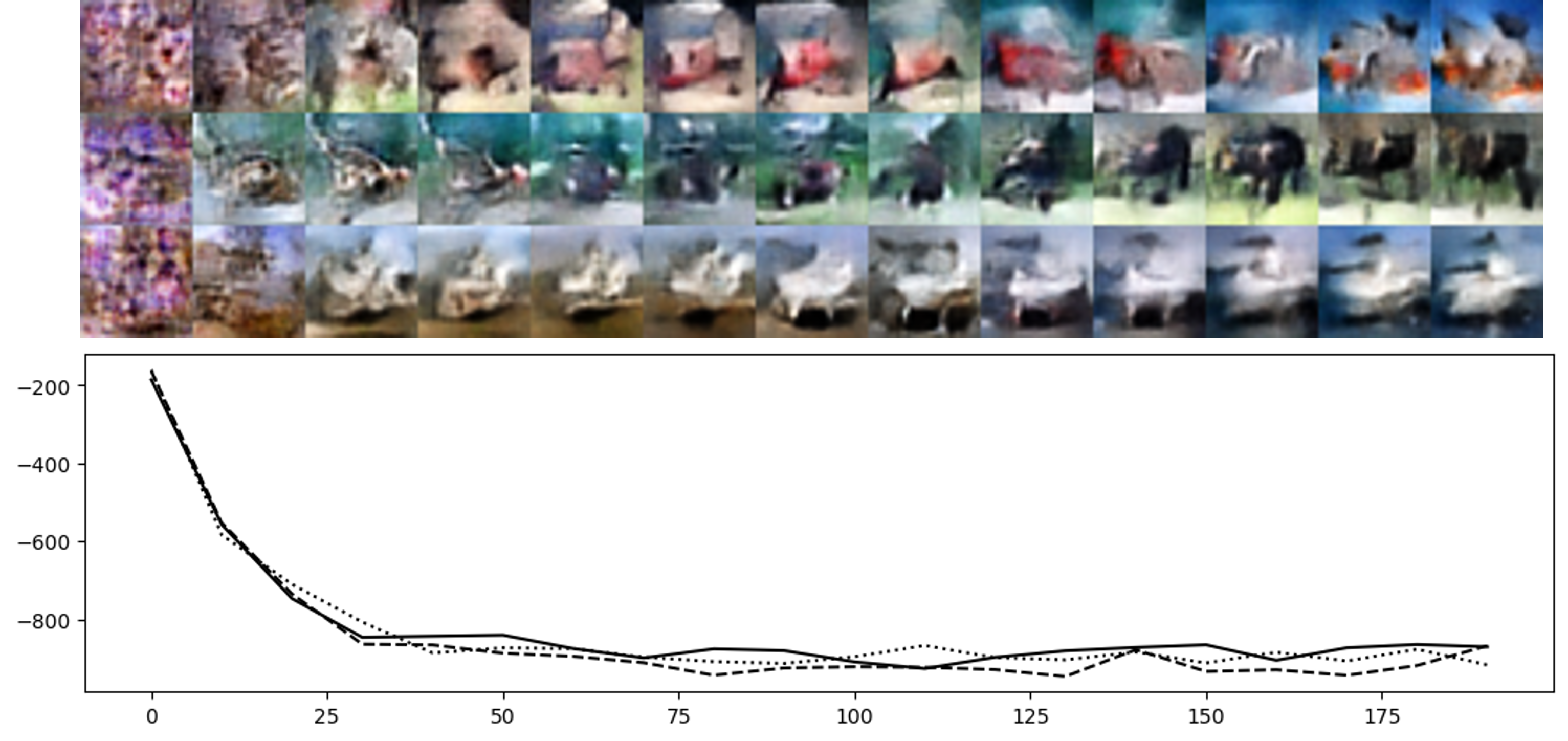}
    }
    \subfigure[2500 steps]{
        \includegraphics[width=0.48\textwidth]{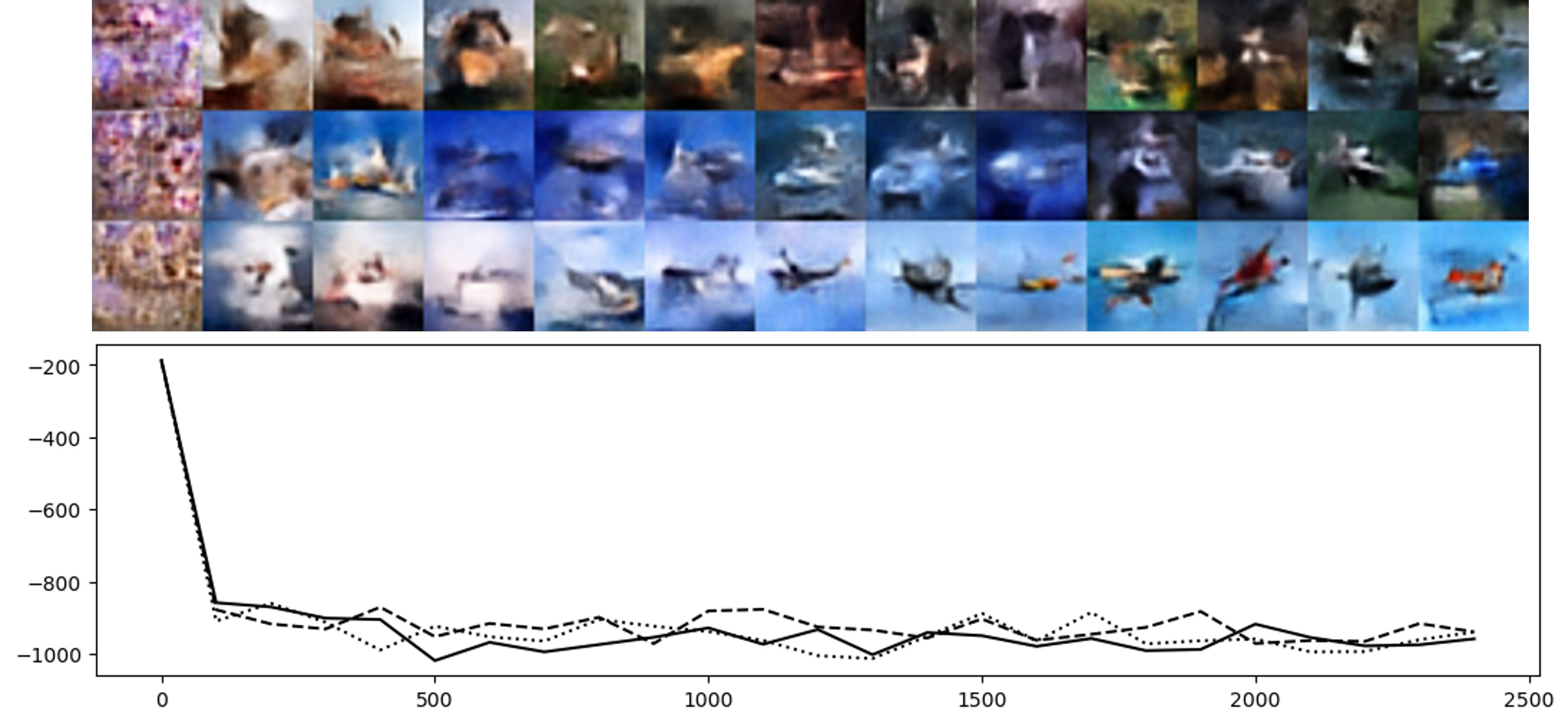}
    }
    \caption{\textbf{Transition of Markov chains initialized from $\N(0, \I_d)$ towards $p_{\bA}(\z)$ on CIFAR-10.} We present results by running \ac{ld} for 200 and 2500 steps. In each sub-figure, the top panel displays the trajectory in the data space uniformly sampled along the chain. The bottom panel shows the energy score $f_{\bA}(\z)$ over the iterations.}
    \label{fig:anl_cifar10}
    \vspace{-10pt}
\end{figure}

We provide additional visualization results of 
\ac{ld} transitions initialized from $\N(0, \I_d)$ on SVHN (\cref{fig:anl_svhn}) and CIFAR-10 datasets (\cref{fig:anl_cifar10}). For the 200-step set-up, we can see that the generation quality quickly improves by exploring the local modes with \ac{ld}. For the 2500-step long-run set-up, we can see that the \ac{ld} produces consistently valid results without the oversaturating issue of the long-run chain samples.

\clearpage
\newpage
\section{Additional Quantitative Results}
\label{app:quati_res}
\subsection{Learning \texorpdfstring{\acs{damc}}{}
in the Latent Space of Other \texorpdfstring{\acp{dlvm}}{}}
We have compared our method with directly training diffusion models in the latent space. In the NALR-LEBM column in \cref{table:abl}, we compare our method with learning a diffusion model in the pre-trained energy-based latent space on CIFAR-10 dataset. We further add experiments of learning this model in the pre-trained VAE \citep{kingma2013auto} and ABP \citep{han2017alternating} model latent spaces summarized in \cref{tab:baselines_app}. We can see that our method greatly outperforms these baselines using the same network architectures.

\begin{table}[!htbp]
    \centering
    \caption{\textbf{Further baseline results for learning \ac{damc} on CIFAR-10 dataset.} These models are implemented using the same encoder and decoder network architectures for fair comparison}
    \begin{tabular}{ccccc}
    \toprule
    {} & VAE & ABP & NALR & Ours \\
    \midrule
    FID  & 102.54 & 69.93	 
         &  64.38 & \textbf{57.52} \\ 
    MSE  & 0.036  & 0.017	 
         & 0.016	 & \textbf{0.015} \\
    \bottomrule
    \end{tabular}
    \label{tab:baselines_app}
\end{table}

\subsection{Further Ablation Studies}
We conduct more ablation studies to inspect the impact of some hyper-parameters. For prior \ac{ld} sampling, we follow \citep{pang2020learning} and set the sampling step as $60$ throughout the experiments for fair comparison. We further add experiments on CIFAR-10 for training our model with different posterior \ac{ld} sampling steps \texttt{T} and with different capacities of the amortizer $q_{\bP}$, summarized in \cref{tabl:abl_app}. We observe that larger posterior sampling steps and larger model capacities for training brings marginal improvement compared with our default set-up. Less sampling steps and lower model capacities, however, may have negative impact on the performances.

\begin{table}[ht]
\centering
 \begin{minipage}[t]{.35\textwidth}
 \centering
 (a) Posterior sampling steps \texttt{T} \\
 \begin{tabular}{cccc}
    \toprule
    {} & \texttt{T=10} & \texttt{T=30} & \texttt{T=50} \\
    \midrule
    FID  & 74.20 & \textbf{57.72} & 57.03 \\ 
    MSE  & 0.016 & \textbf{0.015}	 
         & 0.015 \\
    \bottomrule
 \end{tabular}
 \end{minipage}
 \begin{minipage}[t]{.55\textwidth}
  \centering
  (b) Model capacity of $q_{\bP}$\\
  \begin{tabular}{cccccc}
    \toprule
    {} & \texttt{f=1/4} & \texttt{f=1/2} & \texttt{f=1} &
    \texttt{f=2} & 
    \texttt{f=4} \\
    \midrule
    FID  & 116.28 & 80.28 & \textbf{57.52} & 57.82 & 57.56 \\ 
    MSE  & 0.017 & 0.016 & 
    \textbf{0.015} & 0.015 & 0.015 \\
    \bottomrule
    \end{tabular}
 \end{minipage}
 \caption{\textbf{Ablation Studies for the choice of Langevin steps and model capacity}. We highlight the results of our set-up reported in the main text. \texttt{T} is the posterior sampling step. \texttt{f} stands for the factor for the model capacity, \eg, \texttt{f=2} means 2x the size of the original model.}
 \label{tabl:abl_app} 
\end{table}

\clearpage
\newpage
\section{Further Discussion}
\subsection{Limitations}
\label{app:limit}
We mentioned in the main text that one potential disadvantage of our method is its parameter inefficiency for introducing an extra \acs{ddpm}. Although fortunately, our models are in the latent space so the network is lightweight. To be specific, on SVHN, CelebA, CIFAR-10 and CelebA-HQ datasets the number of parameters in the diffusion network is around $10\%$ of those in the generator.  

Another issue is the time efficiency. We mentioned in the main text that the time efficiency for sampling is competitive. With the batch size of $64$, on these datasets the \acs{damc} prior sampling takes $0.3s$, while $100$ steps of short-run \ac{ld} with \acs{lebm} takes $0.2s$. The \acs{damc} posterior sampling takes $1.0s$, while \acs{lebm} takes $8.0s$. However, during training we need to run 30 steps of posterior \ac{ld} sampling and 60 steps of prior \ac{ld} sampling in each training iteration. We observe that the proposed learning method takes 15.2 minutes per training epoch, while the short-run \ac{ld}-based learning method takes 14.8 minutes per epoch. These methods are slower than the VAE-base method, which takes 5.5 minutes for an training epoch. We can see that the time efficiency for training is generally bottlenecked by the \ac{ld} sampling process, and could be improved in future works.

\subsection{Broader Impacts}
Generative models could be misused for disinformation or faking profiles. Our work focuses on the learning algorithm of energy-based prior model. Though we consider our work to be foundational and not tied to particular applications or deployments, it is possible that more powerful energy-based generative models augmented with this method may be
used maliciously. Work on the reliable detection of synthetic content could be important to address such harms from generative models.

\end{document}